\pdfoutput=1

\documentclass[11pt]{article}

\usepackage{EMNLP2022}

\usepackage{times}
\usepackage{latexsym}

\usepackage[T1]{fontenc}

\usepackage[utf8]{inputenc}

\usepackage{microtype}

\usepackage{inconsolata}

\usepackage{xspace}
\usepackage{amsmath}
\usepackage{subcaption}
\usepackage{tabularx}
\usepackage{booktabs}

\usepackage{graphicx}
\usepackage{enumitem}
\usepackage{csvsimple}
\usepackage{dsfont}
\usepackage{pifont}
\usepackage{amsthm}
\usepackage[ruled,vlined]{algorithm2e}
\usepackage{xspace}
\usepackage{amsmath}
\usepackage{arydshln}
\usepackage{multirow}
\usepackage{colortbl}
\usepackage{float}
\usepackage{multicol}
\usepackage{soul}
\usepackage[colorinlistoftodos,prependcaption,textsize=tiny]{todonotes}
\usepackage[normalem]{ulem}
\usepackage{xcolor}

\usepackage{enumitem}
\setlist{parskip=2pt}
\setlist{topsep=2pt}

\usepackage[T1]{fontenc}

\usepackage[utf8]{inputenc}

\usepackage{microtype}

\newcommand{\scifact}{\textsc{SciFact}\xspace}

\newcommand{\dataname}{\textsc{SciFact-Open}\xspace}
\newcommand{\scifactOpen}{\dataname}
\newcommand{\scifactOrig}{\textsc{SciFact-Orig}\xspace}

\newcommand{\fever}{\textsc{Fever}\xspace}

\newcommand{\storc}{\textsc{S2ORC}\xspace}

\newcommand{\roberta}{RoBERTa\xspace}

\newcommand{\longformer}{Longformer\xspace}
\newcommand{\pjoint}{\textsc{ParagraphJoint}\xspace}
\newcommand{\arsjoint}{\textsc{ARSJoint}\xspace}
\newcommand{\sysname}{\textsc{MultiVerS}\xspace}
\newcommand{\sysnameTen}{$\sysname_{10}$\xspace}

\newcommand{\vertserini}{\textsc{Vert5Erini}\xspace}

\newcommand{\nei}{\textsc{NEI}\xspace}
\newcommand{\support}{\textsc{Support}\xspace}
\newcommand{\refute}{\textsc{Refute}\xspace}
\newcommand{\supports}{\textsc{Supports}\xspace}
\newcommand{\refutes}{\textsc{Refutes}\xspace}
\newcommand{\supported}{\textsc{Supported}\xspace}
\newcommand{\refuted}{\textsc{Refuted}\xspace}

\definecolor{lightblue}{RGB}{212, 235, 255}
\definecolor{salmon}{RGB}{255, 164, 168}
\definecolor{lightgreen}{RGB}{177, 231, 171}
\definecolor{lightyellow}{RGB}{255, 255, 148}
\sethlcolor{lightblue}

\newcolumntype{P}[1]{>{\centering\arraybackslash}p{#1}}
\newcolumntype{L}[1]{>{\raggedright\let\newline\\\arraybackslash\hspace{0pt}}m{#1}}
\newcolumntype{R}[1]{>{\raggedleft\arraybackslash}p{#1}}
\newcommand\tworows[1]{\multirow{2}{*}{\shortstack[l]{#1}}}

\newcounter{djw}

\definecolor{supports}{HTML}{008B72}
\definecolor{refutes}{HTML}{B7321C}
\newcommand{\supportsColor}{\textcolor{supports}{\supports}\xspace}
\newcommand{\refutesColor}{\textcolor{refutes}{\refutes}\xspace}

\newcommand{\cA}{{\mathcal{A}}}

\definecolor{brightmaroon}{rgb}{0.76, 0.23, 0.28}

\setlist{parskip=2pt}
\setlist{topsep=2pt}

\title{\scifactOpen: Towards open-domain scientific claim verification}

\author{David Wadden$^\mathbf{\dagger}$ \quad
  Kyle Lo$^\mathbf{\ddagger}$ \quad
  Bailey Kuehl$^\mathbf{\ddagger}$ \quad
  \textbf{Arman Cohan}$^\mathbf{\ddagger}$ \\
  \textbf{Iz Beltagy}$^\mathbf{\ddagger}$ \quad
  \textbf{Lucy Lu Wang}$^\mathbf{\dagger\ddagger}$ \quad
  \textbf{Hannaneh Hajishirzi}$^\mathbf{\dagger\ddagger}$\\
  $^\mathbf{\dagger}$
  University of Washington, Seattle, WA, USA \\
  $^\mathbf{\ddagger}$
  Allen Institute for Artificial Intelligence, Seattle, WA, USA \\
  {\tt\small \{dwadden,hannaneh\}@cs.washington.edu}, {\tt\small lucylw@uw.edu}, \\
  {\tt\small \{kylel,baileyk,armanc,beltagy\}@allenai.org}
}

\begin{document}
\maketitle

\begin{abstract}
 While research on scientific claim verification has led to the development of powerful systems that appear to approach human performance, these approaches have yet to be tested in a realistic setting against large corpora of scientific literature. Moving to this open-domain evaluation setting, however, poses unique challenges; in particular, it is infeasible to exhaustively annotate all evidence documents. In this work, we present \scifactOpen, a new test collection designed to evaluate the performance of scientific claim verification systems on a corpus of 500K research abstracts. Drawing upon pooling techniques from information retrieval, we collect evidence for scientific claims by pooling and annotating the top predictions of four state-of-the-art scientific claim verification models. We find that systems developed on smaller corpora struggle to generalize to \scifactOpen, exhibiting performance drops of at least 15 F1. In addition, analysis of the evidence in \scifactOpen reveals interesting phenomena likely to appear when claim verification systems are deployed in practice, e.g., cases where the evidence supports only a special case of the claim. Our dataset is available at \url{https://github.com/dwadden/scifact-open}.
\end{abstract}

\section{Introduction}

\begin{figure}[t]
  \centering
  \includegraphics[width=\columnwidth]{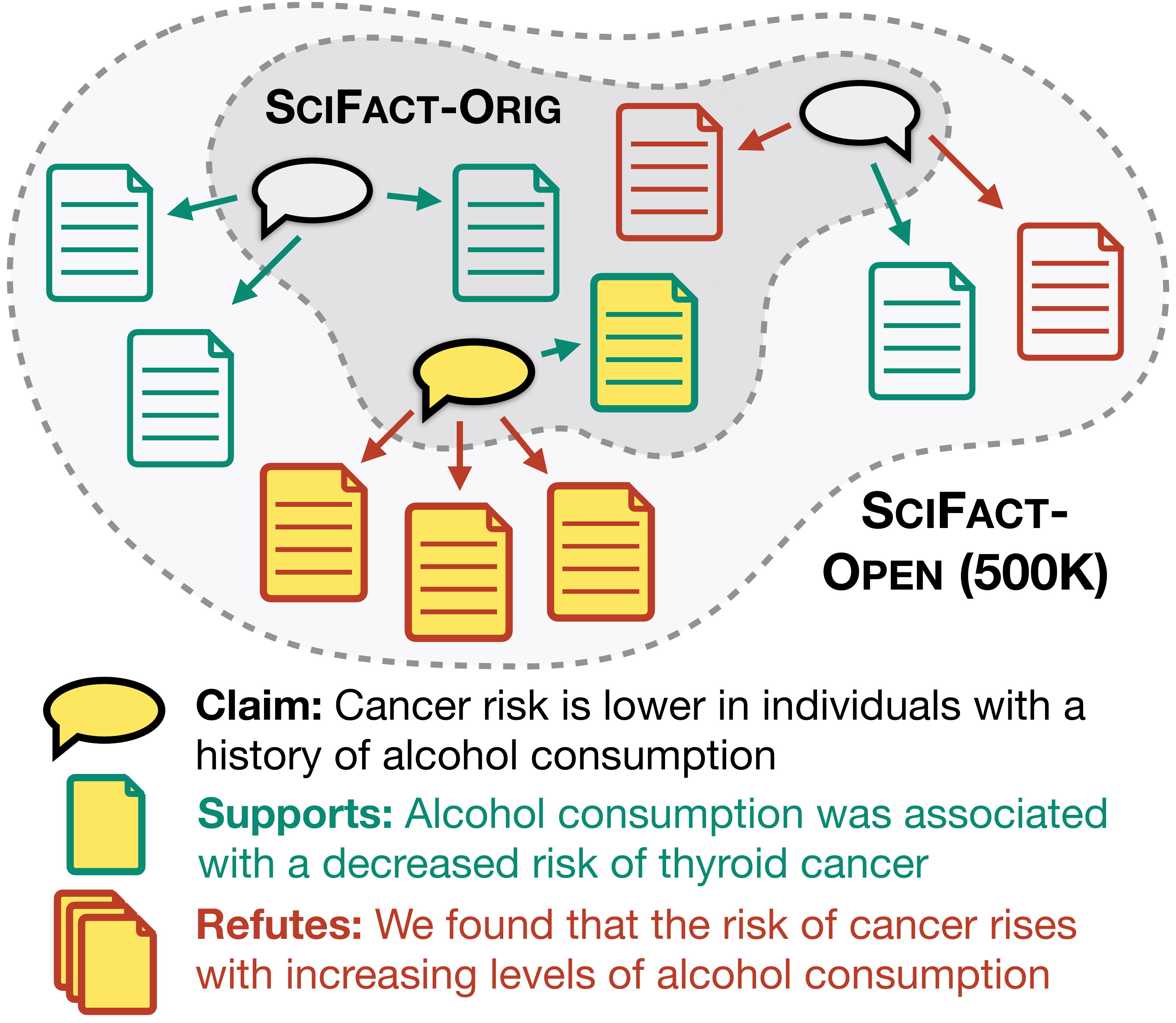}
 \caption{
 \scifactOpen, a new test collection for scientific claim verification that expands beyond the 5K abstract retrieval setting in the original \scifact dataset~\citep{Wadden2020FactOF} to a corpus of 500K abstracts. Each claim in \scifactOpen is annotated with evidence that \supportsColor or \refutesColor the claim. In the example shown, the majority of evidence \refutesColor the claim that alcohol consumption reduces cancer risk, although one abstract indicates that alcohol consumption may reduce thyroid cancer risk specifically.
 }

  \label{fig:teaser}
\end{figure}

The task of scientific claim verification~\cite{Wadden2020FactOF,Kotonya2020ExplainableAF} aims to help system users assess the veracity of a scientific claim relative to a corpus of research literature. Most existing work and available datasets focus on verifying claims against a much more limited context---for instance, a  single article or text snippet \cite{Saakyan2021COVIDFactFE,Sarrouti2021EvidencebasedFO,Kotonya2020ExplainableAF} or a small, artificially-constructed collection of documents \cite{Wadden2020FactOF}. Current state-of-the-art models are able to achieve very strong performance on these datasets, in some cases approaching human agreement \cite{Wadden2021MultiVerSIS}. 

This gives rise to the question of the scalability of scientific claim verification systems to realistic, \textit{open-domain} settings that
involve verifying claims against corpora containing hundreds of thousands of documents.
In these cases, claim verification systems should assist users by identifying and categorizing all available documents that contain evidence supporting or refuting each claim (Fig. \ref{fig:teaser}). 
However, evaluating system performance in this setting is difficult because exhaustive evidence annotation is infeasible, an issue analogous to evaluation challenges in information retrieval (IR).



In this paper, we construct a new test collection for open-domain scientific claim verification, called \scifactOpen, which requires models to verify claims against evidence from both the \scifact \cite{Wadden2020FactOF} collection, as well as additional evidence from a corpus of 500K scientific research abstracts.
To avoid the burden of exhaustive annotation, we take inspiration from the \emph{pooling} strategy \cite{SparckJones1975ReportOT} popularized by the TREC competitions \cite{Voorhees2005TextRetrieval} and combine the predictions of several state-of-the-art scientific claim verification models---for each claim, abstracts that the models identify as likely to \support or \refute the claim are included as candidates for human annotation.

   



Our main contributions and findings are as follows. (1) We introduce \scifactOpen, a new test collection for open-domain scientific claim verification, including 279 claims verified 
against
evidence retrieved from a corpus of 500K abstracts. (2) We find that state-of-the-art models developed for \scifact perform substantially worse (at least 15 F1) in the open-domain setting, 
highlighting the need to improve upon the generalization capabilities of existing systems. (3) We identify and characterize new dataset phenomena that are likely to occur in real-world claim verification settings. These include mismatches between the specificity of a claim and a piece of evidence, and the presence of conflicting evidence (Fig. \ref{fig:teaser}).

With \scifactOpen, we introduce a challenging new test set for scientific claim verification that more closely approximates how the task might be performed in real-word settings. This dataset will allow for further study of claim-evidence phenomena and model generalizability as encountered in open-domain scientific claim verification.



\section{Background and Task Overview} \label{sec:scifact_background} 

We review the scientific claim verification task, and summarize the data collection process and modeling approaches for \scifact, which we build upon in this work. We elect to use the \scifact dataset as our starting point because of the diversity of claims in the dataset and the availability of a number of state-of-the-art models that can be used for pooled data collection. In the following, we refer to the original \scifact dataset as \scifactOrig.

\subsection{Task definition} \label{sec:task_definition}

Given a claim $c$ and a corpus of research abstracts $\cA$, the scientific claim verification task is to identify all abstracts in $\cA$ which contain evidence relevant to $c$, and to predict a label $y(c, a) \in \{ \supports, \refutes \}$ for each evidence abstract. All other abstracts are labeled $y(c, a) = \textrm{NEI}$ (Not Enough Info). We will refer to a single $(c, a)$ pair as a \emph{claim / abstract pair}, or CAP. Any CAP where the abstract $a$ provides evidence for the claim $c$ (either \supports or \refutes) will be called an \emph{evidentiary CAP}, or ECAP. Models are evaluated on their precision, recall, and F1 in identifying and correctly labeling the evidence abstracts associated with each claim in the dataset (or equivalently, in identifying ECAPs).\footnote{The original \scifact task also requires the prediction of rationales justifying each label. Due to the expense of collecting rationale annotations, in this work we do not require rationales; we evaluate using the \emph{abstract-level label-only F1} metric described in \citet{Wadden2020FactOF}.} 

\subsection{\scifactOrig} \label{sec:scifact_orig_dataset}

\begin{figure*}[t!]
  \centering
  \includegraphics[width=\textwidth]{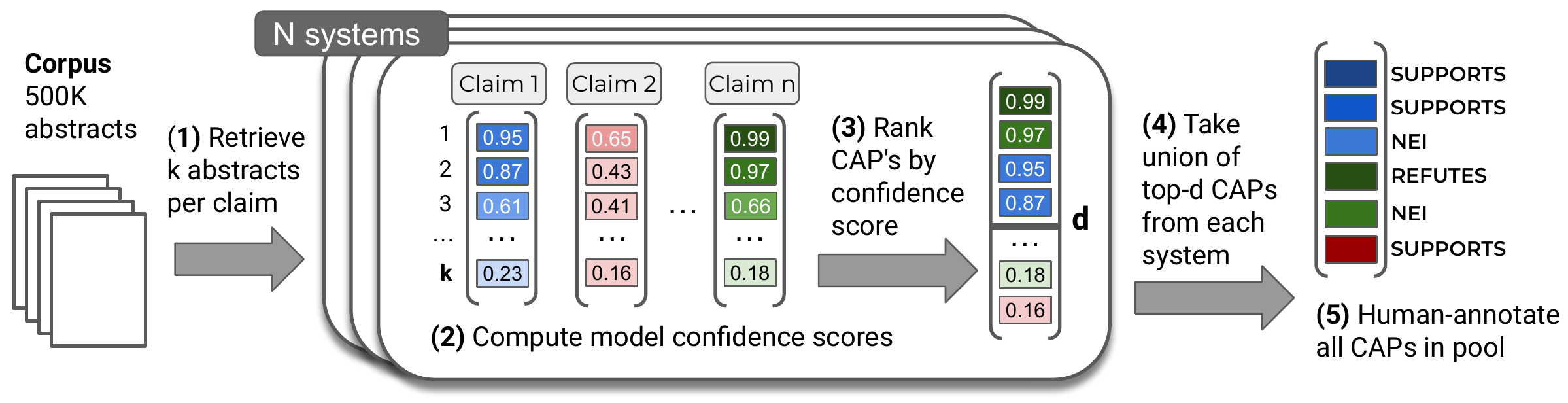}
  \caption{
  Pooling methodology used to collect evidence for \scifactOpen. We construct the pool by combining the $d$ most-confident predictions of $n$ different systems. A single CAP is represented as a colored box; the number in the box indicates a hypothetical confidence score. In this example, the annotation pool contains 3 CAPs from Claim 1, 2 for Claim 2, and 1 for Claim 3. Annotators found evidence for 4 / 6 of these CAPS.
  }
  \label{fig:data_collection}
\end{figure*}

Each claim in \scifactOrig was created by re-writing a citation sentence occurring in a scientific article, and verifying the claim against the abstracts of the cited articles. The resulting claims are diverse both in terms of their subject matter---ranging from molecular biology to public health---as well as their level of specificity (see \S \ref{sec:evidence_phenomena}). Models are required to retrieve and label evidence from a small (roughly 5K abstract) corpus.

Models for \scifactOrig generally follow a two-stage approach to verify a given claim. First, a small collection of candidate abstracts is retrieved from the corpus using a retrieval technique like BM25 \cite{Robertson2009ThePR}; then, a transformer-based language model \cite{Devlin2019BERTPO, Raffel2020ExploringTL} is trained to predict whether each retrieved document \supports, \refutes, or contains no relevant evidence (NEI) with respect to the claim.

As we show in \S \ref{sec:model_performance} and \S \ref{sec:reliability}, a key determinant of system generalization is the \emph{negative sampling ratio}. A negative sampling ratio of $r$ indicates that the model is trained on $r$ irrelevant CAPs for every relevant ECAP. Negative sampling has been shown to improve performance (particularly precision) on \scifactOrig \cite{Li2021APM}. See Appendix \ref{appx:models} for additional details.

\section{The \scifactOpen dataset} \label{sec:data}


In this section, we describe the construction of \scifactOpen. We report the performance of claim verification models on \scifactOpen in \S \ref{sec:model_performance}, and perform reliability checks on the results in \S \ref{sec:reliability}.

Our goal is to construct a test collection which can be used to assess the performance of claim verification systems deployed on a large corpus of scientific literature. This requires a collection of claims, a corpus of abstracts against which to verify them, and evidence annotations with which to evaluate system predictions. We use the claims from the \scifactOrig test set as our claims for \scifactOpen.\footnote{We remove 21 claims (out of 300 total) whose source citations lack important metadata; see Appendix \ref{appx:dataset_construction} for details.} To obtain evidence annotations, we use all evidence from \scifactOrig as evidence in our new dataset and collect additional evidence from the \scifactOpen corpus. 

For our corpus, we filter the \storc dataset \cite{Lo2020S2ORCTS} for all articles which (1) cover topics related to medicine or biology and (2) have at least one inbound and one outbound citation. From the roughly 6.5 million articles that pass these filters, we randomly sample 500K articles to form the corpus for \scifactOpen, making sure to include the 5K abstracts from \scifactOrig. We choose to limit the corpus to 500K abstracts to ensure that we can achieve sufficient annotation coverage of the available evidence. Additional details on corpus construction can be found in Appendix \ref{appx:dataset_construction}.

Unlike \scifactOrig (which is skewed toward highly-cited articles from ``high-impact'' journals), we do not impose any additional quality filters on articles included in \scifactOpen; thus, our corpus captures the full diversity of information likely to be encountered when scientific fact-checking systems are deployed on real-world resources like \storc, arXiv,\footnote{\url{https://arxiv.org}} or PubMed Central.\footnote{\url{https://www.ncbi.nlm.nih.gov/pmc}}

\subsection{Pooling for evidence collection} \label{sec:pooling_methodology}


To collect evidence from the \scifactOpen corpus, we adopt a pooling approach popularized by the TREC competitions:  use a collection of state-of-the-art models to select CAPs for human annotation, and assume that all un-annotated CAPs have $y(c, a) = \nei$. We will examine the degree to which this assumption holds in \S \ref{sec:reliability}.

\paragraph{Pooling approach}

We annotate the $d$ most-confident predicted CAPS from each of $n$ claim verification systems. An overview of the process is in shown in Fig. \ref{fig:data_collection}; we number the annotation steps below to match the figure.

We select the most confident predictions for a single model as follows. \textbf{(1)} For each claim in \scifactOpen, we use an information retrieval system consisting of BM25 followed by a neural re-ranker \cite{Pradeep2021ScientificCV} to retrieve $k$ abstracts from the \scifactOpen corpus. \textbf{(2)} For each CAP, we compute the softmax scores associated with the three possible output labels, denoted $s(\supports), s(\refutes), s(\nei)$. We use $\max(s(\supports), s(\refutes))$ as a measure of the model's confidence that the CAP contains evidence.  \textbf{(3)} We rank all CAPs by model confidence, and add the $d$ top-ranked predictions to the annotation pool. The final pool \textbf{(4)} is the union of the top-$d$ CAPs identified by each system. Since some CAPs are identified by multiple systems, the size of the final annotation pool is less than $n \times d$; we provide statistics in \S \ref{sec:annotation_results}. Finally, \textbf{(5)} all CAPs in the pool are annotated for evidence and assigned a final label by an expert annotator, and the label is double-checked by a second annotator (see Appendix \ref{appx:dataset_construction} for details).

We choose to prioritize CAPS for annotation based on model confidence, rather than annotating a fixed number of CAPs per claim, in order to maximize the amount of evidence likely to be discovered during pooling. In \S \ref{sec:evidence_phenomena}, we confirm that our procedure identifies more evidence for claims that we would expect to be more extensively-studied.


\paragraph{Models and parameter settings}

\begin{table}[t]
    \footnotesize
    \centering
    \begin{tabular}{L{0.33\linewidth}L{0.35\linewidth}R{0.17\linewidth}}
        \toprule
        \tworows{Model} & \tworows{Source}                & Negative sampling \\
        \midrule
        \textbf{Pooling and Eval}                                            \\
        \cmidrule(lr){1-3}
        \vertserini     & \citet{Pradeep2021ScientificCV} & 0                \\
        \pjoint         & \citet{Li2021APM}               & 10               \\
        \sysname        & \citet{Wadden2021MultiVerSIS}   & 20               \\
        \sysnameTen     & \citet{Wadden2021MultiVerSIS}   & 10               \\
        \midrule
        \textbf{Eval only}                                                   \\
        \cmidrule(lr){1-3}
        \arsjoint       & \citet{Zhang2021AbstractRS}     & 12             \\
        \bottomrule
    \end{tabular}

    \caption{
        Models used for pooled data collection and evaluation (top), and for evaluation only (bottom). ``Negative sampling'' indicates the negative sampling ratio. \sysnameTen shares the same architecture as \sysname, but trains on fewer negative samples. 
        }
    \label{tbl:models}
\end{table}

We set $k=50$ for abstract retrieval. In practice, we found that the great majority of evidentiary abstracts were ranked among the top 20 retrievals for their respective claims (Appendix \ref{appx:retrieval}), and thus using a larger $k$ would serve mainly to increase the number of irrelevant results. We set $d=250$; in \S \ref{sec:pool_depth}, we show that this is sufficient to ensure that our dataset can be used for reliable model evaluation. 

For our models, we utilized all state-of-the-art models developed for \scifactOrig for which modeling code and checkpoints were available (to our knowledge). We used $n=4$ systems for pooled data collection. During evaluation, we included a fifth system --- \arsjoint --- which became available after the dataset had been collected. Model names, source publications, and negative sampling ratios are listed in Table \ref{tbl:models}; see Appendix \ref{appx:dataset_construction} for additional details.

\subsection{Dataset statistics} \label{sec:annotation_results}

We summarize key properties of \scifactOpen. Table \ref{tbl:dataset_summary} provides an overview of the claims, corpus, and evidence in the dataset. Table \ref{tbl:precision_by_n_systems} shows the fraction of CAPs annotated during pooling which were judged to be ECAPs (i.e. to contain evidence). Overall, roughly a third of predicted CAPs were judged as relevant; this indicates that existing systems achieve relatively low precision when used in an open-domain setting. Relevance is somewhat higher (roughly 50\%) for CAPs predicted by more than one system. The majority of CAPs are selected by a single system only, indicating high diversity in model predictions. As mentioned in \S \ref{sec:pooling_methodology}, the total number of annotated CAPs is 732 (rather than 4 models $\times$ 250 CAPs / model $= 1000$) due to overlap in system predictions.

Table \ref{tbl:scifact_recovered} shows how many of the ECAPs from \scifactOrig would have been annotated by our pooling procedure. The fact that the great majority of the original ECAPs would have been included in the annotation pool suggests that our approach achieves reasonable evidence coverage.

\begin{table}[t]
    \footnotesize
    \centering

\begin{subtable}{\linewidth}
    \begin{tabular}{
            L{0.12\linewidth}
            L{0.12\linewidth}
            L{0.28\linewidth}
            L{0.13\linewidth}
            L{0.11\linewidth}
        }
        \toprule
               &        & \multicolumn{3}{c}{ECAPs}      \\
        \cmidrule(lr){3-5}
        Claims & Corpus & \scifactOrig & Pooling & Total \\
        \midrule
        279    & 500K   & 209          & 251     & 460   \\
        \bottomrule
    \end{tabular}


    \subcaption{Summary of the \scifactOpen dataset, including the number of claims, abstracts, and ECAPs (evidentiary claim / evidence pairs). ECAPs come from two sources: those from \scifactOrig, and those discovered via pooling.
    }
    \label{tbl:dataset_summary}

\end{subtable}


\begin{subtable}{\linewidth}

    \begin{tabular}{lrrr}
        \toprule
        Num. systems & Annotated & Evidence & \% Evidence \\
        \midrule
        1            & 528       & 154      & 29.2        \\
        2            & 150       & 71       & 47.3        \\
        3            & 44        & 20       & 45.5        \\
        4            & 10        & 6        & 60.0        \\
        \cmidrule(lr){1-4}
        All          & 732       & 251      & 34.3        \\
        \bottomrule
    \end{tabular}

    \subcaption{Relevance of CAPs annotated during the pooling process. The first row indicates that 528 CAPs were identified for pooling by one system only; of those CAPs, 154 were judged by annotators as containing evidence. The more systems identified a given CAP, the more likely it is to contain evidence.}
    \label{tbl:precision_by_n_systems}
\end{subtable}


\begin{subtable}{\linewidth}

    \begin{tabular}{L{0.15\linewidth}P{0.2\linewidth}P{0.2\linewidth}P{0.2\linewidth}}
        \toprule
                   & Total & Retrieved & Annotated \\
        \midrule
        ECAPs       & 209   & 187 (89\%)       & 171 (82\%)       \\
        \bottomrule
    \end{tabular}

    \subcaption{Count of how many ECAPs from \scifactOrig would have been identified during pooled data collection. ``Retrieved'' indicates the number of ECAPs that would have been retrieved among the top $k$, and ``Annotated'' indicates the number that would further have been included in the annotation pool.}
    \label{tbl:scifact_recovered}

\end{subtable}

    \caption{Annotation results and dataset statistics for \scifactOpen.}
    \label{tbl:annotation_results}
\end{table}

\subsection{Evidence phenomena in \scifactOpen} \label{sec:evidence_phenomena}

We observe three properties of evidence in \scifactOpen that have received less attention in the study of scientific claim verification, and that can inform future work on this task.

\paragraph{Unequal allocation of evidence}

\begin{figure}[t]
  \centering
  \includegraphics[width=\columnwidth]{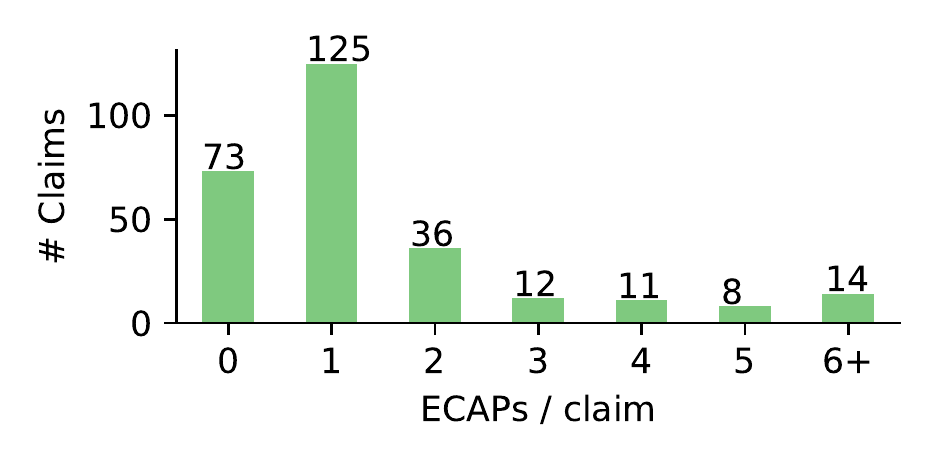}
  \caption{
    Evidence allocation among claims in \scifactOpen. The x-axis indicates the number of ECAPs (evidentiary claim / abstract pairs) associated with a given claim, and the y-axis is the number of claims with the corresponding number of ECAPS. For instance, 125 claims are associated with a single evidence-containing abstract. 
  }

  \label{fig:evidence_distribution}
\end{figure}

Fig. \ref{fig:evidence_distribution} shows the distribution of evidence amongst claims in \scifactOpen. We find that evidence is distributed unequally; half of all ECAPs are allocated to 34 highly-studied claims (12\% of all claims in the dataset). 
We investigated the characteristics of highly-studied claims, and found that they tend to be short and mention a small number of common, well-studied scientific entities. For instance, entities mentioned in well-studied claims ($\geq$ 4 ECAPs) return, on average, 4 times as many documents when entered into a PubMed search, compared to claims with no evidence (detailed results in Appendix \ref{appx:dataset_stats}). Table \ref{tbl:evidence_count_examples_short} shows an example.

\begin{table}[t]
    \footnotesize
    \centering

    \begin{tabular}{
            L{0.8\linewidth} | R{0.1\linewidth}
        }
        \toprule
        \textbf{Claim}                                                                                     & \textbf{ECAPs} \\
        \midrule
        \textbf{Obesity} is determined in part by \textbf{genetic factors}.                                & 19             \\
        \cmidrule(lr){1-2}
        Inhibiting \textbf{HDAC6} decreases survival of \textbf{mice} with \textbf{ARID1A mutated tumors}. & 0              \\
        \bottomrule
    \end{tabular}

    \caption{
        Example of a claim with a number of ECAPs annotated during pooled data collection (top), and another with no new ECAPs (bottom). Well-studied ECAPs tend to be shorter and mention a small number of common entities.
    }
    \label{tbl:evidence_count_examples_short}

\end{table}

\paragraph{Mismatch in claim and evidence specificity} During evidence collection for \scifactOpen, annotators reported situations where a claim and abstract exhibited a relationship, but where the claim applied at a different level of \emph{specificity} from the evidence. For instance, in Fig. \ref{fig:teaser}, the claim and refuting evidence discuss the effects of alcohol consumption on \emph{overall} cancer risk, while the supporting evidence indicates that alcohol consumption lowers thyroid cancer risk \emph{in particular}; the supporting evidence is \emph{more specific} than the claim. We also saw cases where the abstract was \emph{more general} than the claim (e.g. claim discusses thyroid cancer, abstract discusses cancer in general), and where the abstract was \emph{closely related} to the claim (e.g. claim discusses thyroid cancer, abstract discusses throat cancer).\footnote{In cases of mismatching evidence, we follow the convention used in \citet{Thorne2018FEVERAL} to assign an overall \supports\ / \refutes\ / \nei label; see Appendix \ref{appx:partial_evidence} for details.}

Based on this observation, we attempted to quantify the frequency of specificity mismatches. For 206 CAPs in the \scifactOpen annotation pool, in addition to collecting a \supports\ / \refutes\ / \nei label, annotators indicated the specificity relationship between claim and abstract, and wrote a \emph{revision} of the claim such that the revised claim matched the specificity of the abstract. These annotations will be released as part of \scifactOpen.

Table \ref{tbl:partial_evidence_counts} shows counts for different specificity relationships. We find that 91 / 206 (44\%) of the examined CAPs exhibit some form of specificity mismatch. Table \ref{tbl:partial_evidence_example} shows an example where the evidence is more specific than the claim, along with a revised version of the claim that matches the specificity of the evidence. Examples for all specificity relation types --- along with analysis showing that mismatches occur in both well-studied and less-studied claims --- are included in Appendix \ref{appx:partial_evidence}. We discuss possible implications of specificity mismatch for future work on scientific claim verification in \S \ref{sec:discussion}.

\begin{table}[t]
    \footnotesize
    \centering

    \begin{subtable}{\linewidth}
        \centering
        \begin{tabular}{
                L{.7\linewidth}
                R{.1\linewidth}
            }
            \toprule
            \textbf{Category}                 & \textbf{ECAPs} \\
            \midrule
            Evidence matches claim            & 115            \\
            Evidence more specific than claim & 53             \\
            Evidence more general than claim  & 18             \\
            Evidence closely related to claim & 20             \\
            \bottomrule
        \end{tabular}
        \caption{Specificity relationship between claim and evidence, for 206 ECAPs. Specificity mismatches are common, comprising 44\% of annotated examples.}
        \label{tbl:partial_evidence_counts}
    \end{subtable}

    \begin{subtable}{\linewidth}
        \begin{tabular}{
                L{0.95\linewidth}
            }
            \toprule
            \textbf{Claim}: Teaching hospitals provide better care than non-teaching hospitals.                                                                  \\
            \cmidrule(lr){1-1}
            \textbf{Evidence}: Teaching centers \dots prolong survival \emph{in women with any gynecological cancer} compared to community or general hospitals. \\
            \cmidrule(lr){1-1}
            \textbf{Revised claim}: Teaching hospitals provide better \emph{gynecological cancer} care than non-teaching hospitals                               \\
            \bottomrule
        \end{tabular}
        \caption{A CAP where the evidence \supports a special case of the claim, paired with a \emph{revised} version of the claim that matches the evidence. The claim discusses medical care overall, while the evidence discusses gynecological cancer care specifically.}
        \label{tbl:partial_evidence_example}
    \end{subtable}

    \caption{
        Claim / evidence specificity mismatch in \scifactOpen. 
    }
    \label{tbl:partial_evidence}
\end{table}

\begin{table*}[t]
  \footnotesize
  \centering
  \begin{tabular}{
        L{0.16\linewidth}
        *{6}{P{0.05\linewidth}}
        P{0.16\linewidth}
    }
    \toprule

                   & \multicolumn{3}{c}{\textbf{\scifactOrig}} & \multicolumn{4}{c}{\textbf{\scifactOpen}}                                   \\
    \cmidrule(lr){2-4} \cmidrule(lr) {5-8}
    \textbf{Model} & P    & R    & F1                          & P            & R            & F1                    & Average Precision     \\
    \midrule
    \vertserini    & 64.0 & 73.0 & 68.2                        & $25.0_{1.9}$ & $67.2_{2.9}$ & $36.4_{2.2}$          & $27.5_{3.2}$          \\
    \pjoint        & 75.8 & 63.5 & 69.1                        & $54.7_{3.2}$ & $46.5_{3.5}$ & $50.3_{2.7}$          & $40.5_{3.1}$          \\
    \sysname       & 73.8 & 71.2 & 72.5               & $73.6_{2.9}$ & $40.7_{3.3}$ & ${52.4}_{3.0}$ & ${44.9}_{3.7}$ \\
    \sysnameTen    & 63.0 & 73.0 & 67.6                        & $49.6_{3.0}$ & $53.0_{3.7}$ & $51.3_{2.4}$          & $43.4_{3.4}$          \\
    \arsjoint$^*$  & 72.2 & 70.3 & 71.2                        & $46.1_{2.9}$ & $37.6_{3.4}$ & $41.4_{2.7}$          & -                     \\
    \bottomrule
\end{tabular}

  \caption{
    System performance on \scifactOpen. For comparison, metrics on \scifactOrig are also reported. Performance is substantially lower on \scifactOpen relative to \scifactOrig. Precision, recall, and F1 vary widely by system, based on the negative sampling rate used during training. Subscripts indicate standard deviations over 1,000 bootstrap-resampled versions of the claims in \scifactOpen (see Appendix \ref{appx:model_performance}). \\
    $^*$The results for \arsjoint are not comparable with the other systems, since \arsjoint was not used for data collection. We did not compute model confidence scores for \arsjoint; therefore average precision is not reported.
    }
  \label{tbl:metrics}
\end{table*}

\paragraph{Conflicting evidence} Conflicting evidence occurs when a single claim is \supported by at least one ECAP in \scifactOpen, and \refuted by another (see Fig. \ref{fig:teaser}). Of the 81 claims in \scifactOpen with at least 2 ECAPs, 16 of them (20\%) have conflicting evidence. In examining these conflicts, we found that they were often a result of specificity mismatches as shown in Fig. \ref{fig:teaser} (see Appendix \ref{appx:dataset_stats} for additional examples), indicating that modeling evidence specificity represents an important area for future work.

\section{Model performance on \scifactOpen} \label{sec:model_performance}

We evaluate all models from Table \ref{tbl:models} on \scifactOpen. These models represent the state-of-the-art on \scifactOrig, making them strong baselines to assess the difficulty of our new test collection.

\paragraph{\scifactOpen is challenging}   

Table \ref{tbl:metrics} shows the performance of all models on \scifactOpen, as well as on \scifactOrig for comparison. Due to the wide variation in the precision and recall of different models on \scifactOpen, we also report average precision, which summarizes performance via the area under the precision / recall curve. We find that models rank similarly on F1 and average precision. Model performance drops by 15 to 30 F1 on \scifactOpen relative to \scifactOrig, indicating that all models have trouble generalizing to large corpora unseen during training. \pjoint, \sysname, and \sysnameTen all exhibit similar performance (within one standard deviation of each other), while \vertserini performs worse due to low precision. 

In Appendix \ref{appx:model_performance}, we examine model performance on well-studied and less-studied claims separately. We find that higher-recall models tend to perform better on well-studied claims (for which more evidence is available), while higher-precision models perform better on less-studied claims.

\paragraph{Negative sampling affects generalization}

As mentioned in \S \ref{sec:scifact_orig_dataset}, all models except \vertserini were trained with negative sampling. We observe that negative sampling rate has a much larger impact on precision and recall in the open setting than was observed for \scifactOrig. \vertserini has recall more than double its precision; for \sysname, the situation is reversed. The behavior of \sysnameTen is much more similar to \pjoint than \sysname, indicating that negative sampling has a larger impact on model generalization behavior than does model architecture. \arsjoint is qualitatively similar to \pjoint and \sysnameTen, but with lower overall performance since its top predictions are not annotated for evidence (see \S \ref{sec:system_inclusion}).

\begin{figure}[t]
  \centering
  \includegraphics[width=0.8\columnwidth]{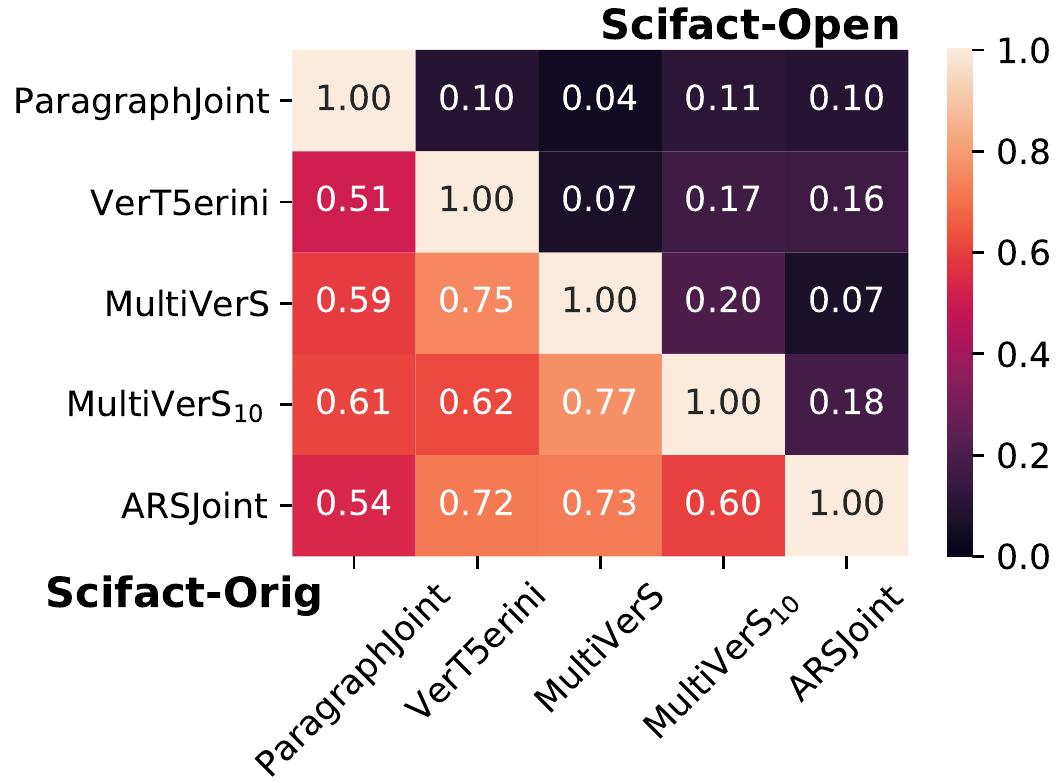}

  \caption{
    Overlap between the ECAPs predicted by different systems, as measured by Jaccard similarity. Cells below the diagonal show the similarity for abstracts contained in \scifactOrig, while cells above the diagonal show similarity for abstracts that were added in \scifactOpen. Overlap is high on abstracts from \scifactOrig, but much lower when models generalize to documents not seen during training.
  } 
  \label{fig:prediction_overlap}
\end{figure}

\paragraph{Models have low agreement on \scifactOpen}
Fig. \ref{fig:prediction_overlap} shows the overlap among the ECAPs predicted by different systems, measured using Jaccard similarity. Overlap is relatively high ($\geq 0.5$) for predictions involving abstracts that were found in \scifactOrig, and is much lower ($\leq 0.2$) on abstracts added in \scifactOpen. From a data collection standpoint, low agreement on \scifactOpen is a benefit, as it ensures that a diverse set of documents was included in the annotation pool. From a modeling standpoint, it suggests that agreement between existing models when deployed on novel corpora is lower than what has previously been observed.   Understanding the differences in the information being identified by each model represents an important direction for future work.

\section{Dataset reliability} \label{sec:reliability}

The total number of annotations collected during pooling (\S \ref{sec:pooling_methodology}) is determined by two parameters: the number of annotations per system $d$, and the number of systems $n$ for which we collect annotations. These parameters must be large enough that increasing them further is unlikely to (1) lead to the discovery of a large number of additional ECAPs or (2) alter the performance metrics of models evaluated on the dataset. Following \citet{Zobel1998HowRA}, we conduct checks to ensure that conditions (1) and (2) hold for our choices of $d$ and $n$.

\subsection{Annotations per system} \label{sec:pool_depth}

\begin{figure}[t!]
  \centering

  \begin{subfigure}[t]{\columnwidth}
    \centering
    \includegraphics[width=\columnwidth]{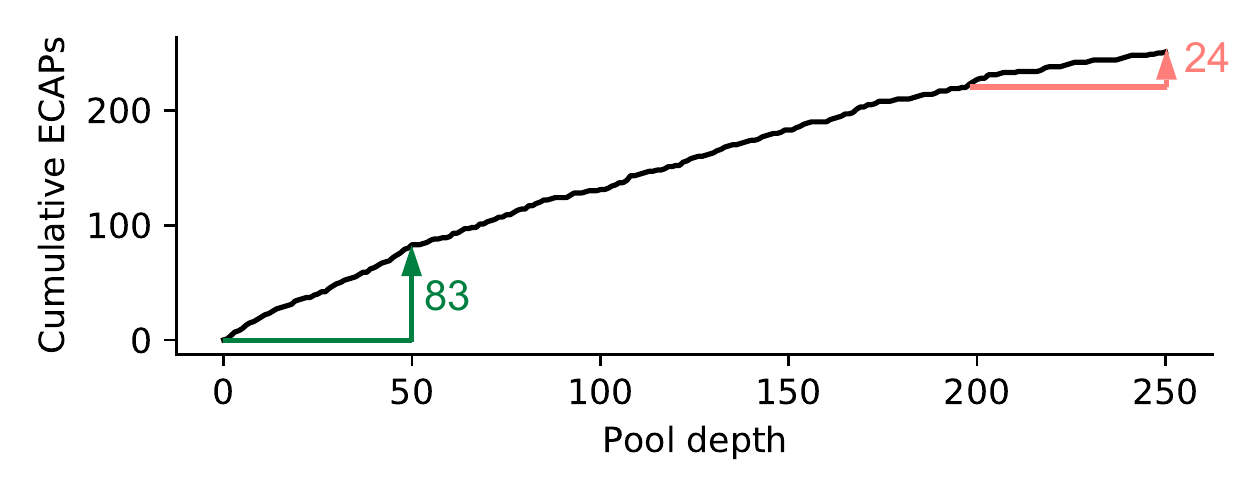}
    \caption{
      Total number of ECAPs discovered as a function of pool depth. For instance, annotating to a depth $d=100$ would have resulted in the discovery of roughly 120 ECAPs. 
    }
    \label{fig:pool_depth_recall}
  \end{subfigure}
  \begin{subfigure}[t]{\columnwidth}
    \centering
    \includegraphics[width=\columnwidth]{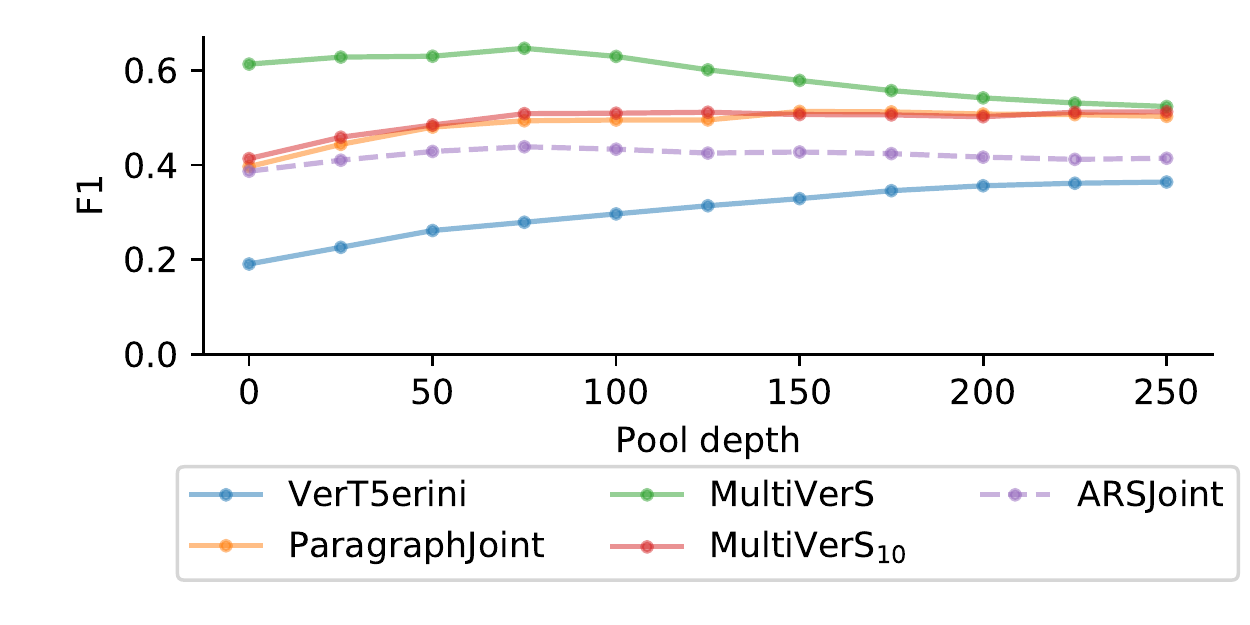}
    \caption{F1 score as a function of pool depth. The blue dot at pool depth 100 indicates that \vertserini would have achieved an F1 score of roughly 30, if annotation had stopped at a depth of 100. Results are \arsjoint are shown as a dashed line to indicate that this system was not used for data collection.
    }
    \label{fig:pool_depth_f1}
  \end{subfigure}


  \caption{
  Effect of pool depth on evidence discovery and evaluation metrics. As pool depth increases, fewer new ECAPs are discovered and F1 score stabilizes.}
  \label{fig:pool_depth}
\end{figure}


To ensure that the number of annotations per system $d=250$ (also called the \emph{pool depth}\footnote{In TREC, the pool depth refers to the number of annotations collected per system for a single query. We use it to refer to the number of annotations collected per system.}) is large enough to ensure reliable evaluation, we examine how much additional evidence is discovered, and how our evaluation metrics change, as $d$ increases from 0 to its final value.

Fig. \ref{fig:pool_depth_recall} shows the total number of ECAPS discovered as a function of pool depth. Annotating the 50 most-confident CAPs per system leads to the discovery of 83 ECAPs, while increasing pool depth from 200 to 250 yields 24 new ECAPs---a more than three-fold decrease. This indicates that condition (1) approximately holds; the majority of the evidence in the corpus has been annotated by $d=250$.

Fig. \ref{fig:pool_depth_f1} shows the F1 score of each model as a function of pool depth. While F1 scores change initially, increasing the pool depth from $d=225$ to $d=250$ changes the F1 score of each model by less than 2\% (see Appendix \ref{appx:reliability} for plots). This indicates that condition (2) also holds: further increases to pool depth are unlikely to affect performance metrics. We also find that generalization behavior is influenced more by negative sampling rate than by model architecture. Performance of \sysname decreases with depth, indicating that it was over-fit to the documents in \scifactOrig, while \vertserini improves with depth. These observations hold if we use average precision rather than F1 to measure performance (Appendix \ref{appx:reliability}).

\subsection{System count} \label{sec:system_count}

\begin{figure}[t!]
  \centering

  \begin{subfigure}[t]{\columnwidth}
    \centering
    \includegraphics[width=\columnwidth]{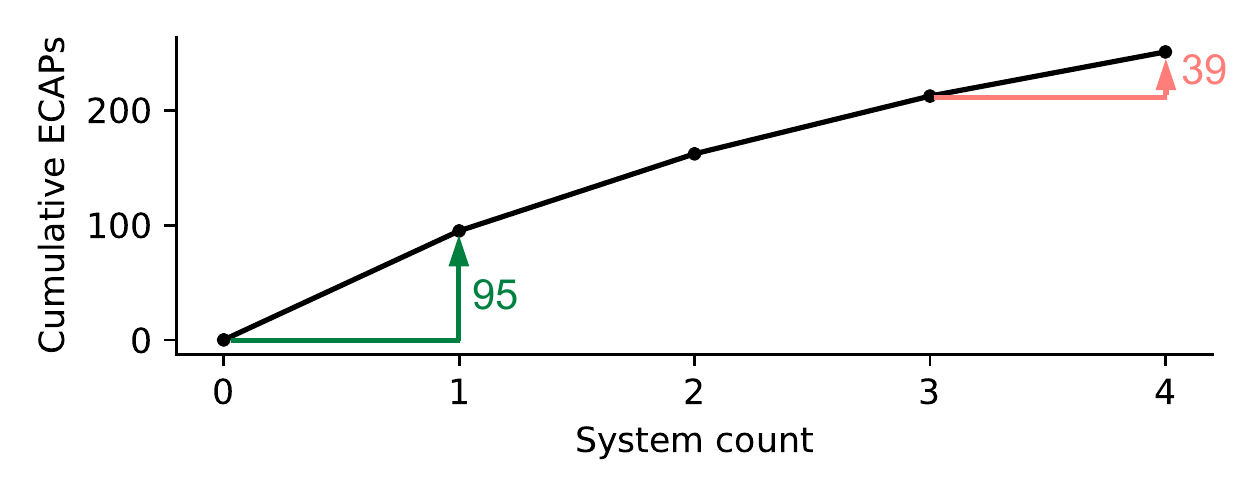}
    \caption{
      Total ECAPs discovered as a function of system count. 
    }
    \label{fig:system_count_recall}
  \end{subfigure}
  
  \begin{subfigure}[t]{\columnwidth}
    \centering
    \includegraphics[width=\columnwidth]{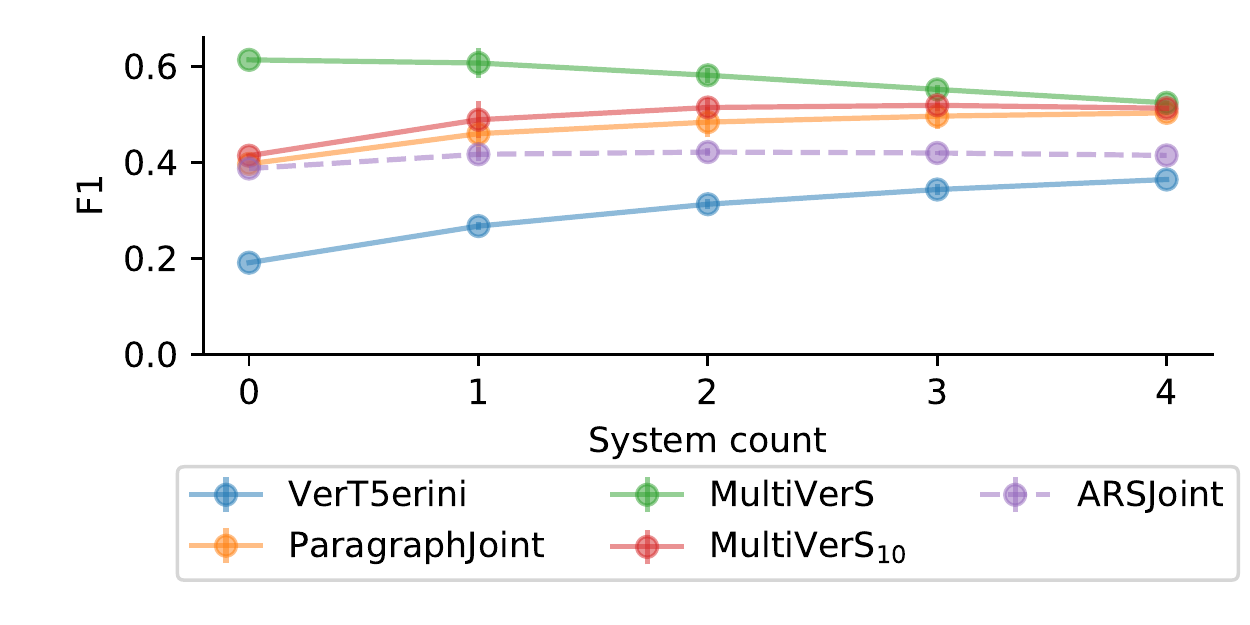}
    \caption{
      F1 score as a function of system count. 
    }
    \label{fig:system_count_f1}
  \end{subfigure}
  
  
  \caption{Effect of system count (i.e. number of systems used during pooling) on evidence discovery and evaluation metrics. As in Fig. \ref{fig:pool_depth}, we see diminishing returns to increasing system count.}
  \label{fig:system_count}
\end{figure}


We repeat the analysis from \S \ref{sec:pool_depth}, but this time varying the number of systems used for data collection (the \emph{system count}).\footnote{There are $4!$ possible system orderings. We compute metrics using all orderings, and display the mean and standard deviation (as error bars) in Fig. \ref{fig:system_count}.} As was the case for pool depth, Fig. \ref{fig:system_count_recall} shows that fewer new ECAPs are discovered as more systems' predictions are annotated. Fig. \ref{fig:system_count_f1} shows that F1 scores stabilize as system count increases, but not as completely as for pool depth; adding a fourth system still leads a 10\% change in F1 score for \vertserini and \sysname (Appendix \ref{appx:reliability}). Thus, while conditions (1) and (2) are increasingly satisfied as the system count increases, \scifactOpen would likely benefit from the collection of additional data identified by new models. Unfortunately, unlike pool depth, the system count that we can achieve is limited by the number of available systems for this task.

\subsection{System inclusion} \label{sec:system_inclusion}

To measure the effect on measured performance of including a given system in the annotation pool, we evaluate each system on the evidence that would have been collected if that system's predictions had not been included. Results are shown in Table \ref{tbl:omission}. All systems except \sysname suffer a roughly 15\% drop. When excluded from data collection, \pjoint and \sysnameTen both have performance comparable to \arsjoint. \sysname does not benefit from having its own predictions included, since it was over-fit to \scifactOrig and struggles to identify new evidence not seen during training. Overall, for fair model comparisons, the performance of new models should be compared against the ``Excluded''  performance of models used for data collection.

\begin{table}
    \footnotesize
    \centering
    \begin{tabular}{lrrr}
        \toprule
        Model          & Included & Excluded & \% Change \\
        \midrule
        \vertserini     & 36.4     & 30.5    & -16.3      \\
        \pjoint & 50.3     & 42.3    & -15.9      \\
        \sysname       & 52.4     & 51.6    & -1.5       \\
        \sysnameTen    & 51.3     & 43.7    & -14.7      \\
        \cmidrule(lr){1-4}
        ARSJoint       & -        & 41.4    & -          \\
        \bottomrule
    \end{tabular}
    \caption{Change in F1 score when each model is included in the annotation pool, vs. excluded. Omission leads to a performance decrease of roughly 15\% for all models except \sysname.}
    \label{tbl:omission}
\end{table}

\section{Related work}

\paragraph{TREC and pooled data collection}

Pooling for IR evaluation was popularized by the TREC information retrieval competitions \cite{Voorhees2005TextRetrieval}, with a number of recent competitions focusing on retrieval in the biomedical domain \cite{Roberts2020TRECCOVIDRA,Roberts2020OverviewOT,Roberts2016OverviewOTC}. Relative to this work, TREC datasets are characterized by a smaller number of queries (or ``topics''), a larger number of models available for annotation, and a fixed number of annotations per topic (often around 50)---although previous works have proposed strategies to prioritize topics or models for annotation \cite{Zobel1998HowRA,Cormack1998EfficientCO}. In contrast, to maximize our annotation yield, we collect a variable number of annotations per claim based on model confidence.

\paragraph{Claim verification and revision}

\citet{Stammbach2021TheCO} studied scientific claim verification against a large research corpus, but simplified the task by evaluating accuracy at predicting a single global truth label per claim, rather than identifying all relevant documents. The Climate-\fever dataset \cite{Diggelmann2020CLIMATEFEVERAD} is also open-domain, but assumes a global truth label and verifies claims against Wikipedia, not research papers.

In \S \ref{sec:evidence_phenomena}, we proposed claim revisions as a solution to claim / evidence specificity mismatch. Claim revision has previously been studied for fact verification over Wikipedia, with the goal of changing the claim from \refuted to \supported or vice versa \cite{Thorne2021EvidencebasedFE,Schuster2021GetYV,Shah2020AutomaticFS}. Previous work has also examined the related task of generating claims based on citation contexts \cite{Wright2022GeneratingSC} and revising questions to match the specificity of answers found in Wikipedia \cite{Min2020AmbigQAAA}.

\section{Discussion \& Conclusion} \label{sec:discussion}

In this work, we introduced a new test collection, \scifactOpen, to support performance evaluation for open-domain scientific claim verification. 
The construction of \scifactOpen was enabled by our adaptation of the pooling strategy from IR for identification and annotation of evidence from a corpus of 500K documents. We hope such methodology can see further usage on other NLP tasks for which exhaustive annotation is infeasible. 


In analyzing the evidence in \scifactOpen (\S \ref{sec:evidence_phenomena}), we found that some claims possess a large amount of conflicting evidence, and that evidence may not always match the specificity of the claims as written. 
We consider two future directions to improve the expressiveness of scientific claim verification. (1) As discussed in \S \ref{sec:evidence_phenomena}, one could still require systems to label each ECAP, but also to generate a revised claim matching the specificity of each evidence abstract. This output would provide users with fine-grained information indicating the conditions under which an input claim is likely to hold. We release 91 claim revisions, which can be used to facilitate exploratory research in this direction. (2) One could use the evidence identified by a claim verification system as input into a summarization system \cite{DeYoung2021MS2MS,Wallace2020GeneratingN} --- potentially using additional quality criteria (e.g. citation count, publication venue) to filter or re-weight the articles included in the summary. This approach has the benefit of providing a concise summary to the user, but there is a greater risk of hallucination \cite{Maynez2020OnFA}.

Overall, our analysis indicates that evaluations using \scifactOpen 
can provide key insights into modeling challenges associated with 
scientific claim verification. In particular, the performance of existing models 
declines substantially
when evaluated on \scifactOpen, suggesting that current claim verification systems are not yet ready for deployment at scale. 
It is our hope that the dataset and analyses presented in this work will facilitate future modeling improvements, and lead to substantial new understanding of the scientific claim verification task.  

\section{Limitations}

A major challenge in information retrieval is the infeasibility of exhaustive relevance annotation. By introducing an open-domain claim verification task, we are faced with similar challenges around annotation. We adopt TREC-style pooling in our setting with substantially fewer systems than what is typically pooled in TREC competitions, which may lead to greater uncertainty in our test collection. We perform substantial analysis (\S \ref{sec:reliability}) to better understand the sensitivity of our test collection to annotation depth and system count, and our results suggest that though further improvements are possible, \scifactOpen is still useful as a test collection and is able to discern substantive performance differences across models. As other models are developed for claim verification, we may indeed incorporate their predictions in pooling to produce a better test collection.

Through analysis of claim-evidence pairs in \scifactOpen, we identified the phenomenon of unequal allocation of evidence (\S \ref{sec:evidence_phenomena}). Some claims are associated with substantially higher numbers of relevant evidence documents; we call these highly-studied claims. In this work, we do not treat these claims any differently than those associated with limited evidence. It could be that highly-studied claims are more representative of the types of claims that users want to verify, in which case we may want to distinguish between these and other types of claims in our dataset, or develop annotation pipelines that would allow us to identify and verify more of these highly-studied claims. In the context of this paper, we derive all claims from the original \scifact test collection, and do not provide additional claims. 


Finally, we rely on a single retrieval system to identify candidate abstracts. While our analysis indicates that this system identifies the great majority of relevant abstracts (Appendix \ref{appx:retrieval}), future work could extend the dataset collected here by retrieving documents using a wider variety of IR approaches.

\section*{Acknowledgments}

This research was supported by NSF IIS-2044660, ONR N00014-18-1-2826, ONR MURI N00014-18-1-2670, a Sloan fellowship and gifts from AI2. We thank the Semantic Scholar team at AI2, UW-NLP, and the H2lab at UW for helpful comments and feedback. Thanks to Xiangci Li and Ronak Pradeep for help with \pjoint and \vertserini, respectively.

\bibliography{refs}
\bibliographystyle{acl_natbib}

\appendix
\section{Dataset construction}  \label{appx:dataset_construction}

\subsection{Corpus}

We use \texttt{2019-09-28} release of the the \storc corpus \cite{Lo2020S2ORCTS} as our source of abstracts. We filter for documents whose \texttt{mag\_field\_of\_study} field includes at least one of \texttt{Medicine} or \texttt{Biology}, and which have at least one inbound and one outbound citation; the latter check serves as a basic quality filter to make sure that the article is related to other articles found in the literature. This filtering leaves us with roughly 6.5 million documents, from which we randomly sample our \scifactOpen corpus of 500K documents (making sure to include the 5K documents from \scifactOrig). For more information on \storc, see \url{https://github.com/allenai/s2orc}.

\subsection{Claims and evidence}

\paragraph{Claims} As mentioned in \S \ref{sec:data}, we removed 21 claims from the \scifactOrig test set when constructing \scifactOpen, leaving 279 claims. Each claim in the dataset is based on a source citation. We removed claims for which we could not find metadata in \storc providing information about the source citation --- in particular, the article it came from or the year the article was published. No analyses based on this information were included in the final version of this work, but the dataset had already been collected when we realized that this claim filtering step had been unnecessary. Regardless, there is no reason that the availability (or lack thereof) of source metadata would correlate with any linguistic properties of the claims; thus, the omission of these 21 claims should not bias our findings in any way.

\paragraph{Evidence annotation}
Evidence annotations were performed by three professional annotators with undergraduate degrees in fields related to biology. Before beginning annotations, they participated in a training session with one of the authors to ensure that they understood the task. Annotations are performed as follows. First, every CAP in the annotation pool is assigned randomly to one of the three annotators. The assigned annotator makes a decision on whether the instance clearly does not contain evidence, or whether it might. If it clearly does not, it is marked \nei and annotation stops. If it might contain evidence, the first annotator assigns a label, and a second annotator checks the label to confirm. In the case of disagreement, the two annotators discuss the instance and collectively decide on a final label.


\subsection{Retrieval} \label{appx:retrieval}

\begin{figure}[t!]
  \centering

  \includegraphics[width=\columnwidth]{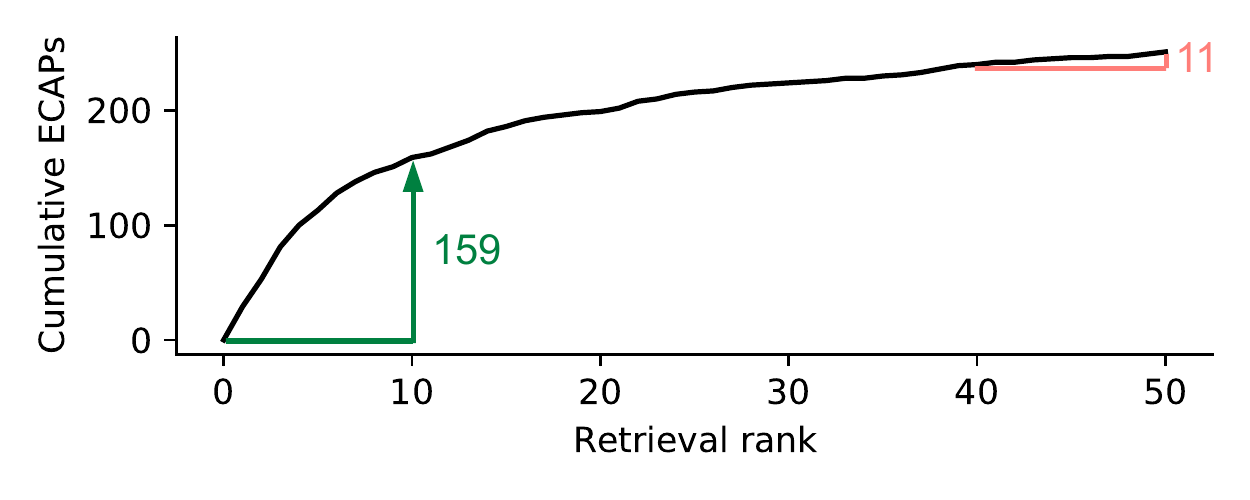}
  \caption{Number of ECAPs discovered as a function of $k$, the number of abstracts retrieved per claim. The great majority of abstracts judged as ECAPs were ranked among the top 20 retrievals for their respective claims.}
  \label{fig:retrieval_depth}
\end{figure}

In \S \ref{sec:pooling_methodology}, we chose to retrieve $k=50$ abstracts per claim; these abstracts were then rank-ordered by model confidence, and the most-confident predictions were annotated for evidence. Using $k=50$ is justified if worse-ranked retrievals are unlikely to be annotated as ECAPs during the pooling process. Figure \ref{fig:retrieval_depth} confirms that this is indeed the case; the great majority of evidentiary abstracts have retrieval ranks of 20 or better, and very few have ranks worse than 40.

\subsection{Models} \label{appx:models}

For pooled annotation collection, we used all models achieving state-of-the-art or competitive performance on the \scifact leaderboard\footnote{\url{https://leaderboard.allenai.org/scifact}} for which modeling code and checkpoints were available as of early summer 2021, when annotation collection began. The available systems were \vertserini \cite{Pradeep2021ScientificCV} and \pjoint \cite{Li2021APM} --- the two leaders on the \texttt{SciVer} shared task \cite{Wadden2021OverviewAI} --- and \sysname \cite{Wadden2021MultiVerSIS}, formerly called \texttt{LongChecker}. Early in annotation, we noticed that the systems exhibited different precision and recall behavior, and hypothesized that this was due to differences in negative sampling rate. To test this, we also collected annotations with a version of \sysname trained with a negative sampling ratio of 10 (negative sampling ratio is defined in \S \ref{sec:scifact_orig_dataset}), referred to as \sysnameTen, and found that this model indeed behaved more like \pjoint than \sysname in terms of precision and recall. We decided to include \sysnameTen in the annotation process to increase the diversity of annotation pool. Subsequently, \arsjoint \cite{Zhang2021AbstractRS} was released and achieved comparable performance with the four systems used for data collection. We conduct evaluations on this system as well.

\paragraph{System descriptions}

Given a claim $c$, all models first retrieve a collection of candidate abstracts $a$, and then predict labels for each retrieved candidate. In this work, we used the \vertserini retrieval system for all models, since it outperformed the performance of the techniques used with \arsjoint and \pjoint. \vertserini first retrieves documents using BM25, then re-ranks the retrieved documents using a neural re-ranker trained on MS-MARCO \cite{Campos2016MSMA}. We experimented with using dense retrieval instead \cite{Karpukhin2020DensePR}, but found that this did not perform well; similar results were reported in \citet{Thakur2021BEIRAH}.

Given a claim $c$ and abstract $a$, \vertserini selects rationales (evidentiary sentences) from $a$ using a T5-3B model trained on \scifact, and then makes label predictions based on the selected rationales using a separate T5-3B model.

\pjoint and \arsjoint both encode the claim and full abstract using \roberta \cite{Liu2019RoBERTaAR}, truncating to 512 tokens, and use these representations as the basis for both rationale selection and label prediction. Rationales are predicted based on self-attention over the encodings of the tokens in each sentence, and then a final label is predicted based on self-attention over the representations of the sentences that were selected as rationales.

\sysname encodes the claim and full abstracts in the same fashion as \pjoint and \arsjoint, using \longformer \cite{Beltagy2020LongformerTL} to accommodate long abstracts, and then predicts the label and rationales in a multitask fashion, based on encodings of the leading \texttt{[SEP]} token and sentence separator tokens, respectively.

\paragraph{Negative sampling}
For scientific claim verification, negative sampling has been performed as follows: for every $(c, a)$ instance in the training data where $y(c, a) \in \{ \supports, \refutes \}$, include $r$ additional $(c, a'_i)_{i=1}^r$ instances where $y(c, a'_i) = \nei$ for all $i$. The irrelevant abstracts $a'_i$ can be sampled randomly from the corpus, or can be chosen to be ``hard'' negatives; for instance, abstracts $a'_i$ could be chosen which have high lexical overlap with claim $c$, but which are not annotated as \supports or \refutes. Negative sampling has been shown to increase the precision of fact verification models \cite{Li2021APM}, but comes at the cost of increasing the size of the training dataset (and thus the training time) by a factor of $r$.


\section{Additional evidence properties}  \label{appx:dataset_stats}

\subsection{Unequal allocation of evidence}

\begin{figure}[t!]
  \centering

  \begin{subfigure}[t]{\columnwidth}
    \centering
    \includegraphics[width=\columnwidth]{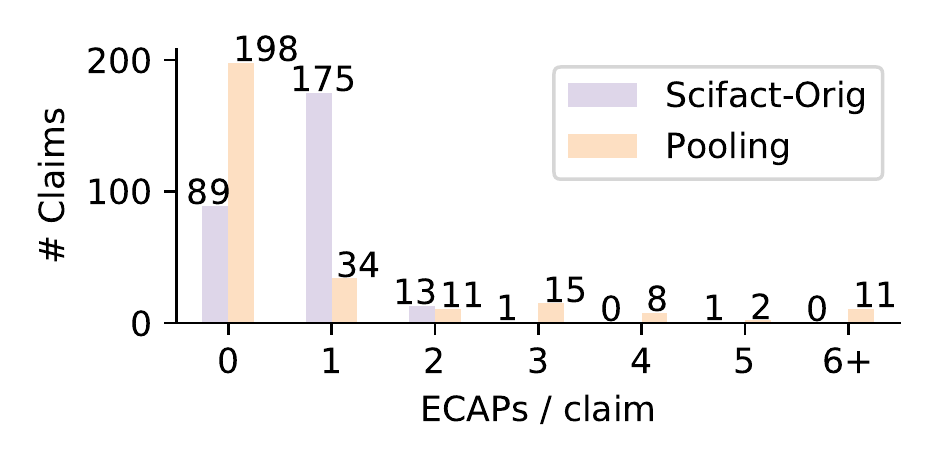}
    \caption{
    Evidence allocation among claims, showing evidence from \scifactOrig and evidence collected during pooling separately. The x-axis indicates the number of ECAPs associated with a given claim, and the y-axis is the number of claims with that number of ECAPS.
    }
    \label{fig:evidence_bar_appx}
  \end{subfigure}

  \begin{subfigure}[t]{\columnwidth}
    \centering
    \includegraphics[width=\columnwidth]{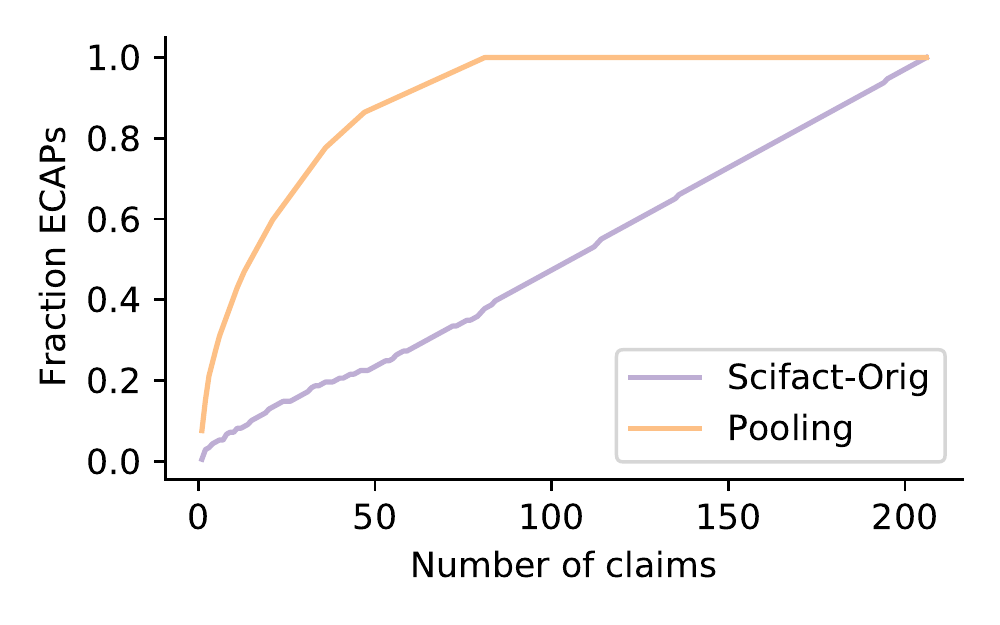}
    \caption{Cumulative distribution of ECAPs across claims.
    }
    \label{fig:evidence_cdf}
  \end{subfigure}
  \caption{
  Distribution of evidence from \scifactOrig, and from the evidence collected via pooling.
  }
  \label{fig:evidence_distribution_appx}
\end{figure}

Fig. \ref{fig:evidence_bar_appx} shows the distribution of evidence amongst claims in \scifactOpen, showing evidence from \scifactOrig and evidence collected during pooling separately. The majority of claims in \scifactOrig have one ECAP. Pooling discovered no new ECAPs for the majority of claims in the dataset, and discovered a large amount of evidence for a small handful of claims. Fig. \ref{fig:evidence_cdf} shows the cumulative distribution of evidence. 14 claims account for 50\% of the ECAPs discovered via pooling.

In \S \ref{sec:evidence_phenomena}, we also observed that well-studied claims tend to be short, and mention a small number of well-studied entities. Table \ref{tbl:claim_evidence_alloc} shows these results quantitatively by examining the characteristics of claims for which pooling discovered at least 4 new ECAPs, vs. claims for which it discovered none. Entities mentioned in well-studied claims return, on average, 4 times as many documents when entered into a PubMed search compared with entities mentioned in claims with no new evidence. We use  BERN2~\cite{Sung2022BERN2AA} to identify the entities for this analysis.

\begin{table}[t]
    \footnotesize
    \centering

    \begin{tabular}{L{0.48\linewidth}R{0.15\linewidth}R{0.2\linewidth}}
        \toprule
                                   & 0 ECAPs & $\geq 4$ ECAPs \\
        \midrule
        Avg. claim length (tokens) & 14.1    & 10.5           \\
        \cmidrule(lr){1-3}
        Avg. entities / claim      & 2.3     & 1.7            \\
        \cmidrule(lr){1-3}
        Median PubMed results      & 49K     & 198K           \\
        \bottomrule
    \end{tabular}

    \caption{
        Characteristics of claims for which 0 ECAPs were annotated during pooled data collection, compared to claims with $\geq 4$ ECAPs annotated. All differences are significant at the $0.05$ level.
    }
    \label{tbl:claim_evidence_alloc}

\end{table}

\subsection{Claim / evidence specificity mismatch} \label{appx:partial_evidence}

\paragraph{Annotation conventions for mismatched evidence}

In situations where the evidence in abstract $a$ is more specific or more general than claim $c$, we follow the convention established in the FEVER dataset \cite{Thorne2018FEVERAL} to assign a final label $y(c, a)$:

\begin{itemize}[leftmargin=*,noitemsep]
    \item If $a$ \supports a special case of $c$, then assign $y(c, a) = \supports$.
    \item If $a$ \supports a generalization of $c$, then assign $y(c, a) = \nei$.
    \item If $a$ \refutes a special case of $c$, then assign $y(c, a) = \nei$.
    \item If $a$ \refutes a generalization of $c$, then assign $y(c, a) = \refutes$.
\end{itemize}

\paragraph{Occurrence for well-studied and less-studied claims}

Table \ref{tbl:partial_evidence_by_popularity} shows rates of claim / evidence specificity mismatch for well-studied and less-studied claims, respectively. Specificity mismatches occur for both types of claims. Interestingly, CAPs where the evidence is more general than the claim occur more frequently for less-studied claims; this likely occurs because less-studied claims are themselves likely to be very specific and cover narrower topics.

\begin{table}[t]
    \footnotesize
    \centering

    \centering
    \begin{tabular}{
            L{.6\linewidth}
            R{.1\linewidth}
            R{.1\linewidth}
        }
        \toprule
        \tworows{\textbf{Category}}       & \multicolumn{2}{c}{\textbf{ECAPs}} \\
                                          & Well-studied & Less-studied        \\
        \midrule
        Evidence matches claim            & 81           & 34                  \\
        Evidence more specific than claim & 35           & 18                  \\
        Evidence more general than claim  & 3            & 15                  \\
        Evidence closely related to claim & 6            & 14                  \\
        \bottomrule
    \end{tabular}
    \caption{Rates of claim / evidence specificity mismatch for well-studied and less-studied claims.}
    \label{tbl:partial_evidence_by_popularity}
\end{table}

\paragraph{Examples}

In \S \ref{sec:evidence_phenomena}, we described how the claim and evidence in an ECAP may not have matching levels of specificity. Table \ref{tbl:partial_evidence_full} provides examples of the different forms of specificity mismatch shown in Table \ref{tbl:partial_evidence}.

\begin{table*}[t]
    \footnotesize
    \centering

    \begin{tabular}{
            p{0.15\linewidth} | p{0.8\linewidth}
        }
        \toprule
        \textbf{Category}    & Evidence \textbf{matches} claim                                                                                                                                                                                                                                        \\
        \arrayrulecolor{black!30}\cmidrule(lr){1-2}
        \textbf{Claim}       & Mitochondria play a major role in calcium homeostasis.                                                                                                                                                                                                                 \\
        \arrayrulecolor{black!30}\cmidrule(lr){1-2}
        \textbf{Evidence}    & Mitochondria \dots are essential organelles responsible for \dots calcium homeostasis.                                                                                                                                                                                 \\
        \arrayrulecolor{black!30}\cmidrule(lr){1-2}
        \textbf{Revision}    &                                                                                                                                                                                                                                                                        \\
        \arrayrulecolor{black!30}\cmidrule(lr){1-2}
        \textbf{Explanation} & No revision necessary; claim and evidence are paraphrases                                                                                                                                                                                                              \\
        \arrayrulecolor{black}\midrule
        \textbf{Category}    & Evidence \textbf{more specific} than claim                                                                                                                                                                                                                             \\
        \arrayrulecolor{black!30}\cmidrule(lr){1-2}
        \textbf{Claim}       & Teaching hospitals provide better care than non-teaching hospitals                                                                                                                                                                                                     \\
        \arrayrulecolor{black!30}\cmidrule(lr){1-2}
        \textbf{Evidence}    & Teaching centres \dots prolong survival \emph{in women with any gynecological cancer} compared to community or general hospitals.                                                                                                                                      \\
        \arrayrulecolor{black!30}\cmidrule(lr){1-2}
        \textbf{Revision}    & Teaching hospitals provide better \emph{gynecological cancer care} than non-teaching hospitals.                                                                                                                                                                        \\
        \arrayrulecolor{black!30}\cmidrule(lr){1-2}
        \textbf{Explanation} & The evidence refers to \emph{gynecological cancer care} specifically, not care care in general.                                                                                                                                                                        \\
        \arrayrulecolor{black}\midrule
        \textbf{Category}    & Evidence \textbf{more general} than claim                                                                                                                                                                                                                              \\
        \arrayrulecolor{black!30}\cmidrule(lr){1-2}
        \textbf{Claim}       & \emph{Somatic missense} mutations in NT5C2 are associated with relapse of \emph{acute lymphoblastic leukemia}.                                                                                                                                                         \\
        \arrayrulecolor{black!30}\cmidrule(lr){1-2}
        \textbf{Evidence}    & T5C2 mutant proteins show \dots resistance to chemotherapy                                                                                                                                                                                                             \\
        \arrayrulecolor{black!30}\cmidrule(lr){1-2}
        \textbf{Revision}    & \emph{Mutations} in NT5C2 are associated with relapse of \emph{cancer}.                                                                                                                                                                                                \\
        \arrayrulecolor{black!30}\cmidrule(lr){1-2}
        \textbf{Explanation} & Evidence mentions T5C2 mutations in general, while the claim mentions \emph{somatic missense mutations} specifically. The evidence discusses chemotherapy resistance generally, while the claim discusses relapse of \emph{acute lymphoblastic leukemia} specifically. \\
        \arrayrulecolor{black}\midrule
        \textbf{Category}    & Evidence \textbf{closely related} to claim                                                                                                                                                                                                                             \\
        \arrayrulecolor{black!30}\cmidrule(lr){1-2}
        \textbf{Claim}       & Near-infrared wavelengths increase penetration depth in \emph{fiberoptic confocal microscopy}                                                                                                                                                                          \\
        \arrayrulecolor{black!30}\cmidrule(lr){1-2}
        \textbf{Evidence}    & Longer wavelength can \dots increase the effective penetration depth of \emph{OCT (optical coherence-domain tomography)} imaging                                                                                                                                       \\
        \arrayrulecolor{black!30}\cmidrule(lr){1-2}
        \textbf{Revision}    & Near-infrared wavelengths increase penetration depth in \emph{optical coherence-domain tomography}.                                                                                                                                                                    \\
        \arrayrulecolor{black!30}\cmidrule(lr){1-2}
        \textbf{Explanation} & The claim discusses \emph{fiberoptic confocal microscopy}. The evidence discusses a different imaging technique, \emph{optical coherence-domain tomography}.                                                                                                           \\
        \arrayrulecolor{black}
        \bottomrule
    \end{tabular}

    \caption{
        Examples of different forms of claim-evidence specificity mismatch. In each example, information specific to claim or evidence is shown in \emph{italics}. The revision re-writes the claim to match the specificity of the evidence.
    }
    \label{tbl:partial_evidence_full}
\end{table*}

\subsection{Conflicting evidence}

Table \ref{tbl:conflicting_evidence} shows examples of two claims for which conflicting evidence was found in \scifactOpen.

\begin{table*}[t]
    \footnotesize
    \centering

    \begin{tabular}{
            p{0.15\linewidth} | p{0.8\linewidth}
        }
        \toprule
        \textbf{Claim}       & Bariatric surgery has a deleterious impact on mental health.                                                                                                           \\
        \cmidrule(lr){1-2}
        \textbf{\supports}   & Our study shows that undergoing bariatric surgery is associated with increases in self-harm, psychiatric service use and occurrence of mental disorders.               \\
        \cmidrule(lr){1-2}
        \textbf{\refutes}    & Statistical analysis revealed significant improvements in depressive symptoms, physical dimension of quality of life, and self-esteem \dots                            \\
        \midrule
        \textbf{Claim}       & Teaching hospitals provide better care than non-teaching hospitals                                                                                                     \\
        \cmidrule(lr){1-2}
        \textbf{\supports}   & Teaching centres or regional cancer centres may prolong survival in women with any gynaecological cancer compared to community or general hospitals                    \\
        \cmidrule(lr){1-2}
        \textbf{\refutes}    & Overall [the results] do not suggest that a healthcare facility's teaching status on its own markedly improves or worsens patient outcomes.                            \\
        \bottomrule
    \end{tabular}

    \caption{
        Examples of two claims with conflicting evidence. 
        }
    \label{tbl:conflicting_evidence}
\end{table*}

\section{Model performance} \label{appx:model_performance}

\paragraph{Uncertainty estimates} Table \ref{tbl:metrics} includes uncertainty estimates for performance on \scifactOpen. We obtain these estimates by computing the standard deviation over 1,000 bootstrap-sampled versions of the dataset \cite{Dror2018TheHG,BergKirkpatrick2012AnEI}. For a single bootstrap iteration, we resample the claims from the dataset with replacement, and evaluate against the evidence for the sampled claims, weighting the evidence by the number of times each claim was sampled.

\paragraph{Performance for well-studied and less-studied claims}

Table \ref{tbl:metrics_by_popularity} shows model performance on well-studied vs. less-studied claims. Higher-recall models tend to perform better on the well-studied claims, since these are the claims where evidence is available in the corpus. Higher-precision models perform better on less-studied claims.  

\begin{table*}[t]
  \footnotesize
  \centering
  \begin{tabular}{
        L{0.16\linewidth}
        *{3}{P{0.05\linewidth}}
        P{0.16\linewidth}
        *{3}{P{0.05\linewidth}}
        P{0.16\linewidth}
    }
    \toprule

                   & \multicolumn{4}{c}{\textbf{Well-studied}} & \multicolumn{4}{c}{\textbf{Less-studied}} \\
    \cmidrule(lr){2-5} \cmidrule(lr) {6-9}
    \textbf{Model} & P    & R    & F1   & Avg. Precision       & P    & R    & F1   & Avg. Precision       \\
    \midrule
    \vertserini    & 31.7 & 65.2 & 42.7 & 30.0                 & 20.9 & 69.1 & 32.1 & 25.5                 \\
    \pjoint        & 57.0 & 39.6 & 46.8 & 40.1                 & 53.2 & 53.2 & 53.2 & 43.8                 \\
    \sysname       & 76.1 & 22.5 & 34.7 & 33.6                 & 72.7 & 58.4 & 64.8 & 56.9                 \\
    \sysnameTen    & 59.3 & 42.3 & 49.4 & 38.6                 & 44.8 & 63.5 & 52.6 & 52.7                 \\
    \arsjoint$^*$  & 37.5 & 22.5 & 28.1 & 11.3                 & 51.0 & 52.4 & 51.7 & 27.0                 \\
    \bottomrule
\end{tabular}

  \caption{Performance on \scifactOpen, for well-studied and less-studied claims. Higher-recall models generally perform better on well-studied claims, while high-precision models perform better on less-studied claims. \\
  $^*$The results for \arsjoint are not comparable with the other systems, since \arsjoint was not used for data collection.
  }
  \label{tbl:metrics_by_popularity}
\end{table*}

\paragraph{Confusion matrices}

\begin{figure*}[t]
    \centering
    \includegraphics[width=0.9\linewidth]{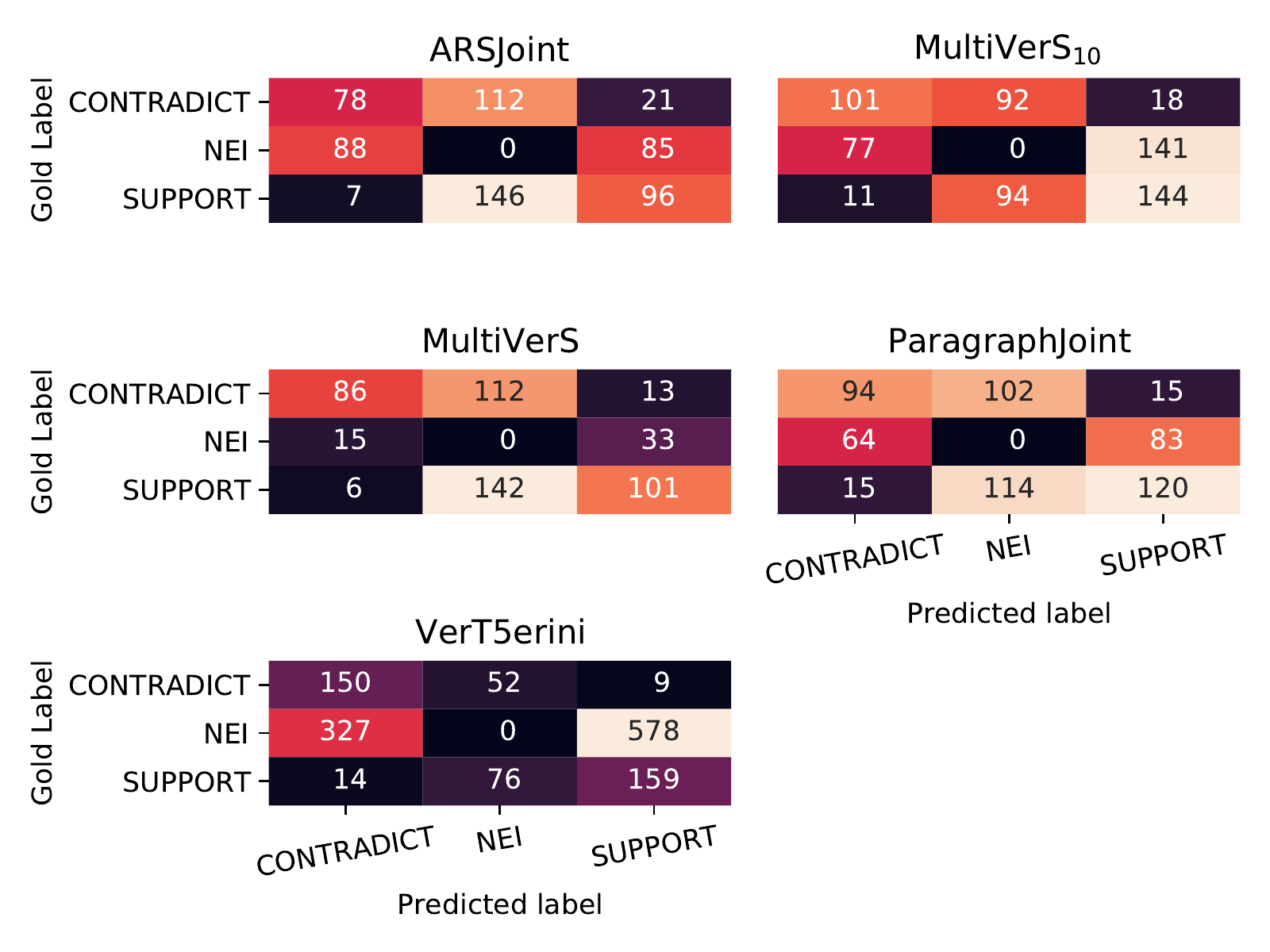}
    \caption{Confusion matrices for model predictions on \scifactOpen. The \{\nei, \nei\} cell is 0 because \scifactOpen is a retrieval task; it's not informative to compute agreement on unlabeled and unpredicted documents in the corpus.}
    \label{fig:model_confusion}
\end{figure*}

Figure \ref{fig:model_confusion} shows confusion matrices for all systems. Models rarely confuse \supports with \refutes; much more commonly, they either mistake irrelevant abstracts for evidence or fail to identify relevant abstracts.

\section{Dataset reliability: Additional experiments} \label{appx:reliability}

\paragraph{Percentage changes in evaluation metrics}

\begin{figure}[t]
  \centering

  \begin{subfigure}[t]{\columnwidth}
    \centering
    \includegraphics[width=\columnwidth]{fig/annotation_depth_f1_True.pdf}
    \caption{F1 score as a function of pool depth. This is the same plot as shown in \S \ref{sec:pool_depth}
    }
    \label{fig:pool_depth_f1_appx}
  \end{subfigure}

  \begin{subfigure}[t]{\columnwidth}
    \centering
    \includegraphics[width=\columnwidth]{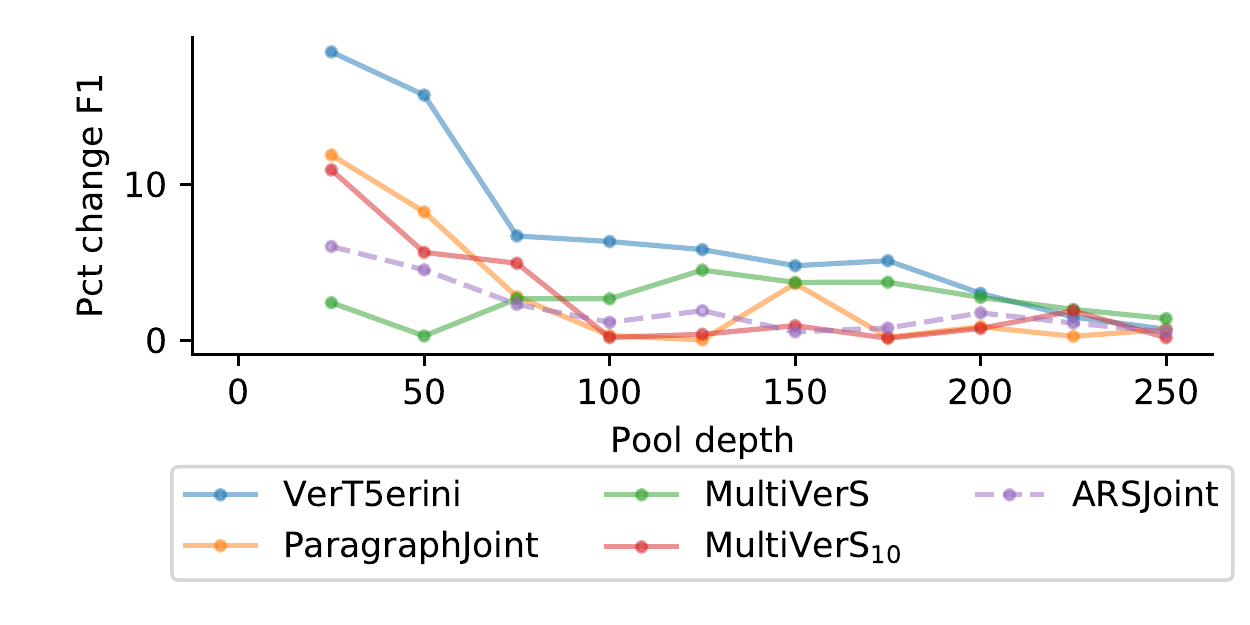}
    \caption{
      Absolute value of percentage change in F1 score, between adjacent points in the series shown in Fig. \ref{fig:pool_depth_f1_appx}. The blue dot at pool depth 100 indicates that the F1 score for \vertserini is increased by roughly 6\% when the pool depth increases from 75 to 100. Values close to 0 indicate that collecting additional data does not have an appreciable effect on F1 score.
      }
    \label{fig:pool_depth_pct_appx}
  \end{subfigure}

  \caption{
  Percentage change in F1 as a function of pool depth.}
  \label{fig:pool_depth_appx}
\end{figure}

\begin{figure}[t]
  \centering

  \begin{subfigure}[t]{\columnwidth}
    \centering
    \includegraphics[width=\columnwidth]{fig/model_depth_f1.pdf}
    \caption{
      F1 score as a function of system count. 
    }
    \label{fig:system_count_f1_appx}
  \end{subfigure}
  
  \begin{subfigure}[t]{\columnwidth}
    \centering
    \includegraphics[width=\columnwidth]{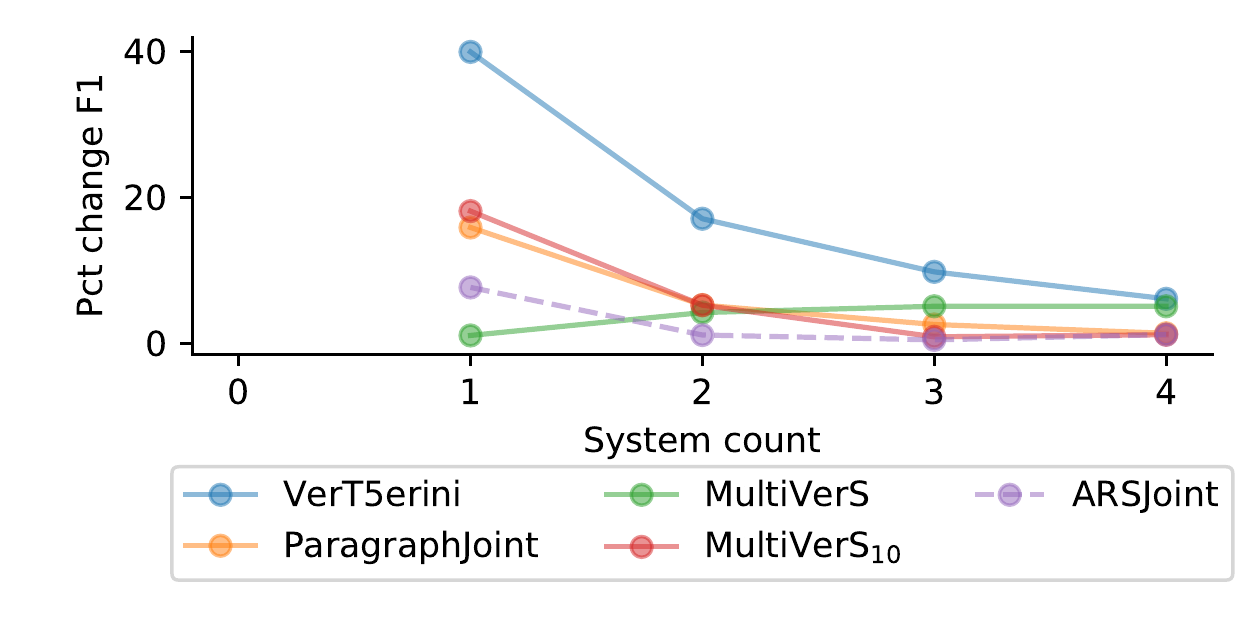}
    \caption{
      Absolute value of percentage change in F1 score, between adjacent points in Fig. \ref{fig:system_count_f1_appx}.
    }
    \label{fig:system_count_pct_appx}
  \end{subfigure}
  
  \caption{Percentage change in F1 as a function of system count.}
  \label{fig:system_count_appx}
\end{figure}

In \S \ref{sec:reliability}, we examined the effect of pool depth and system count on F1 score. Here, we show the same plots from \S \ref{sec:reliability}, together with plots showing the \emph{percentage changes} in the F1 score. Results for pool depth are shown in Fig. \ref{fig:pool_depth_appx}. Results for system count are shown in Fig. \ref{fig:system_count_appx}.

\paragraph{Evaluation using average precision}

\begin{figure}[t!]
  \centering

  \begin{subfigure}[t]{\columnwidth}
    \centering
    \includegraphics[width=\columnwidth]{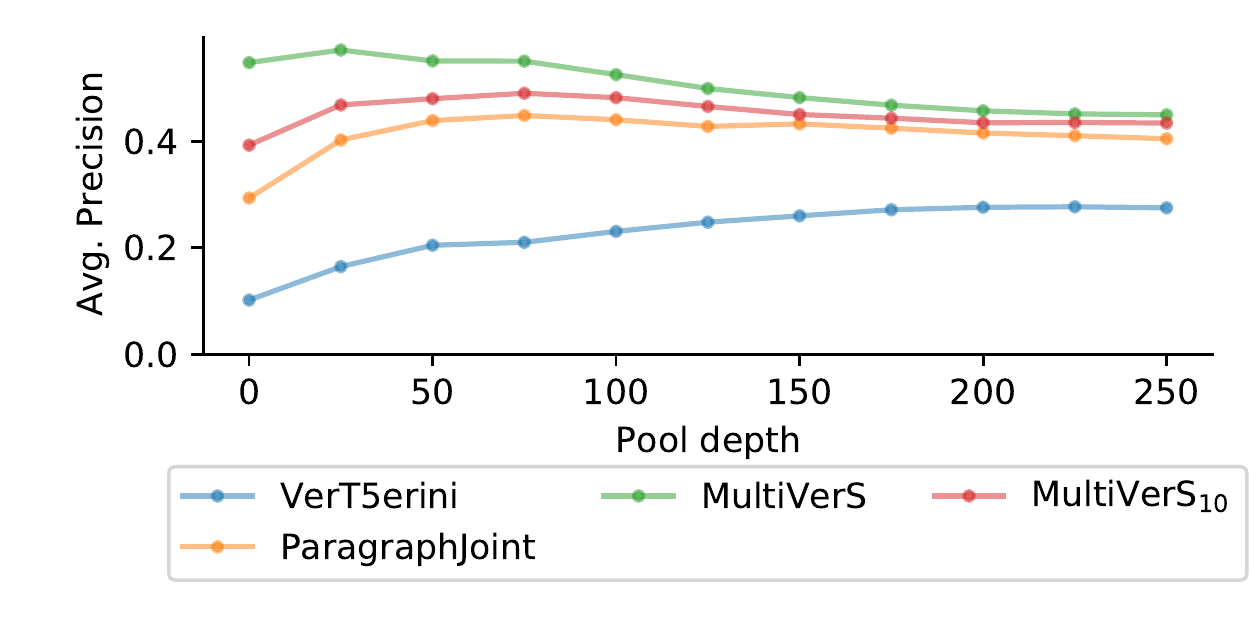}
    \caption{Average precision as a function of pool depth.
    }
  \end{subfigure}
  \begin{subfigure}[t]{\columnwidth}
    \centering
    \includegraphics[width=\columnwidth]{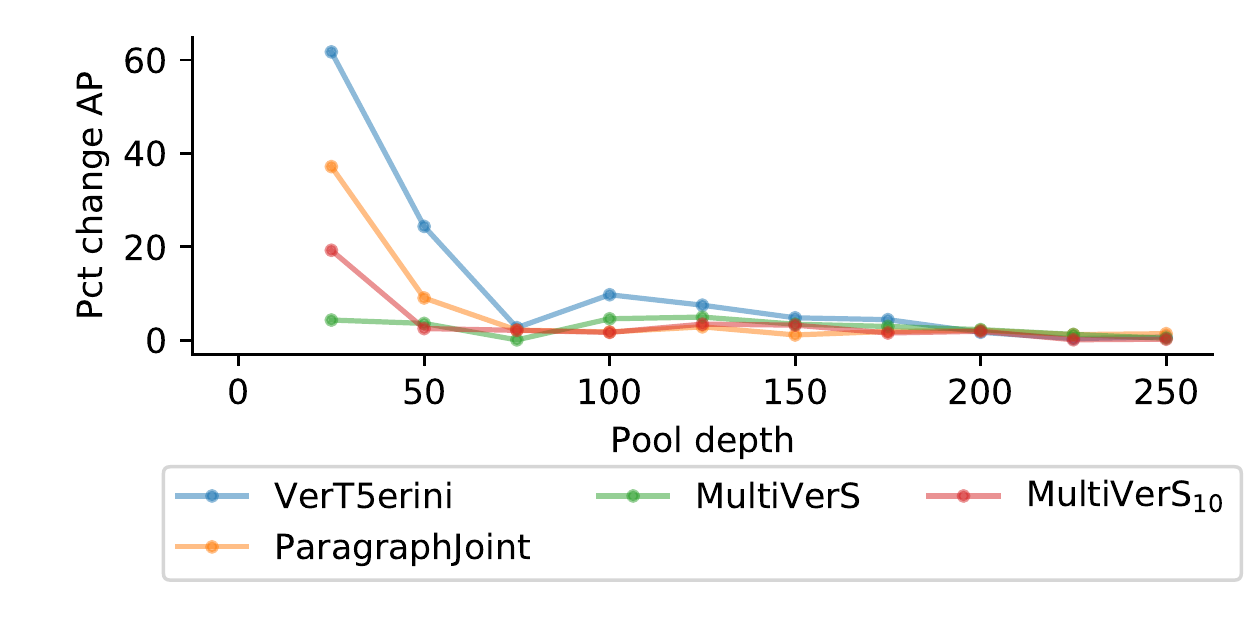}
    \caption{
      Absolute value of percentage change in average precision.
      }
  \end{subfigure}
  \caption{Effect of pool depth on model performance, as measured by average precision. We see the same trends as for F1 score. \arsjoint is not included because computing average precision requires model confidence scores.}
  \label{fig:pool_depth_ap_appx}
\end{figure}

\begin{figure}[t!]
  \centering

  \begin{subfigure}[t]{\columnwidth}
    \centering
    \includegraphics[width=\columnwidth]{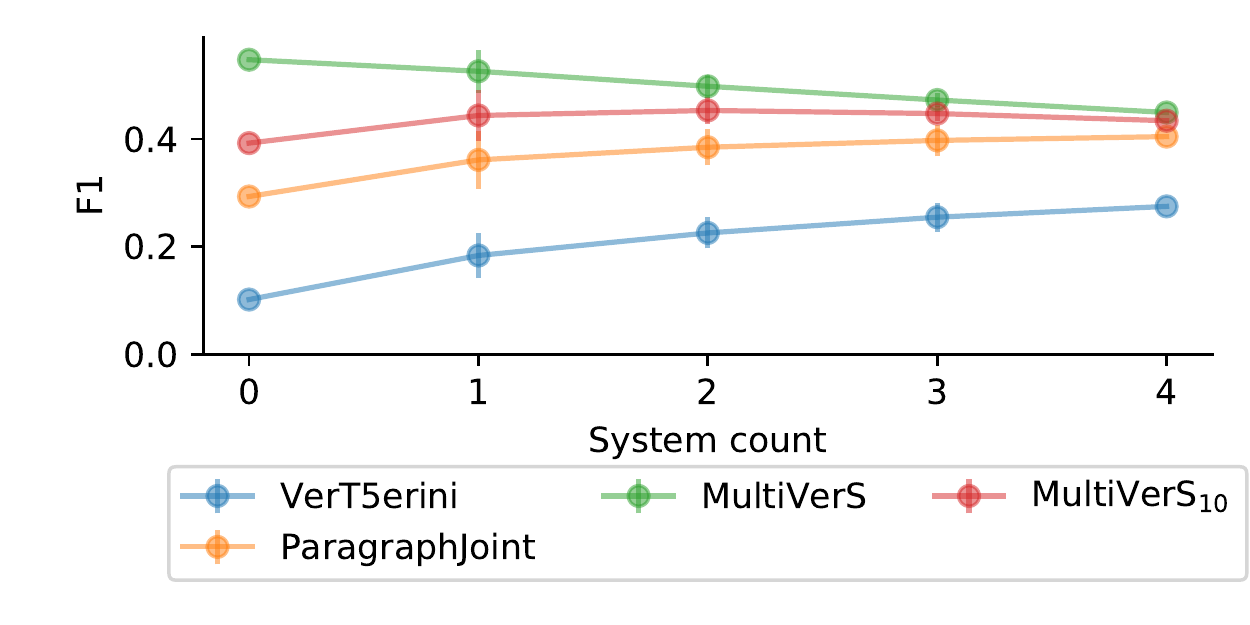}
    \caption{Average precision as a function of system count.
    }
  \end{subfigure}
  \begin{subfigure}[t]{\columnwidth}
    \centering
    \includegraphics[width=\columnwidth]{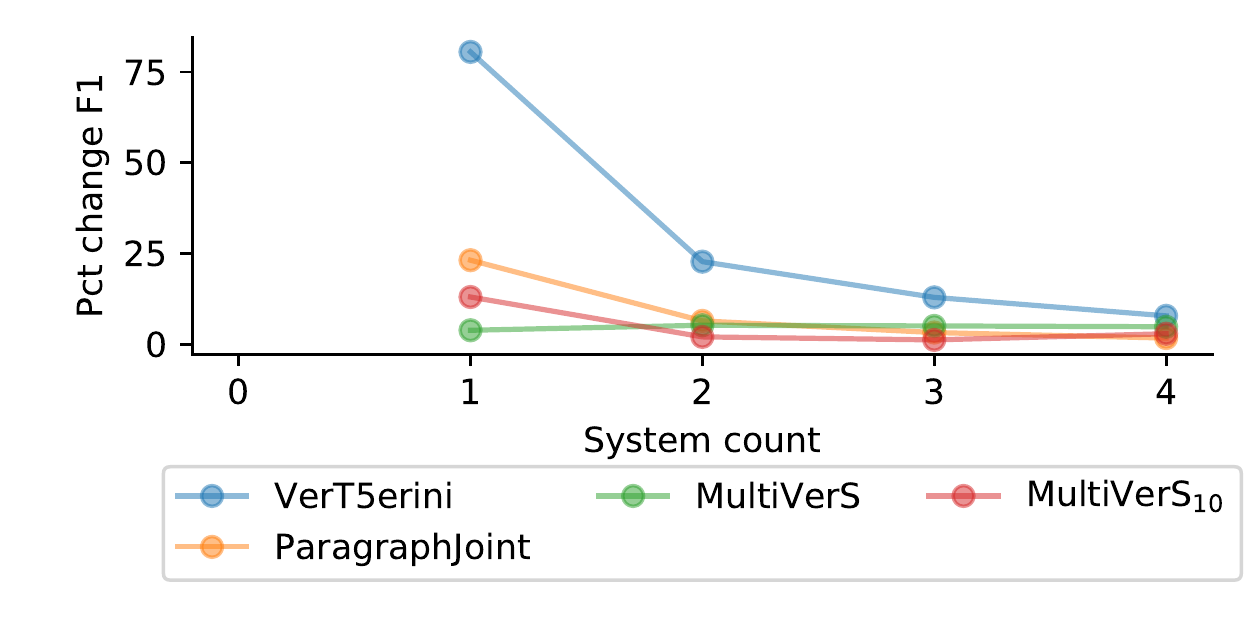}
    \caption{
  Absolute value of percentage change in average precision.
      }
  \end{subfigure}
  \caption{
  Effect of system count on average precision. Again, the results here mirror the conclusions drawn using F1 score.
  }
  \label{fig:model_depth_ap}
\end{figure}

In \S \ref{sec:reliability}, we examined the effect of pool depth and system count on F1 score. We perform the same analysis using average precision. Fig. \ref{fig:pool_depth_ap_appx} shows the effect of pool depth, and Fig. \ref{fig:model_depth_ap} shows the effect of model count. The qualitative conclusions are the same as for F1. The fact that using F1 and average precision leads to the same conclusions indicates that simply re-calibrating each model's classification threshold to adjust for the negative sampling rate used during training would not change the results.

\end{document}